\documentclass{article} 
\usepackage[final]{colm2026_conference}

\usepackage{microtype}
\usepackage{hyperref}
\usepackage{url}
\usepackage{booktabs}

\usepackage{lineno}

\usepackage{tcolorbox}
\tcbuselibrary{breakable}
\usepackage{listings}

\newtcolorbox{promptbox}[1][]{
  fonttitle=\bfseries,
  title=#1,
  breakable,
  boxrule=0.5pt,
  left=6pt,
  right=6pt,
  top=6pt,
  bottom=6pt
}

\usepackage{makecell}
\usepackage{xcolor}
\definecolor{humanblue}{HTML}{2E86AB}
\definecolor{gpt5purple}{HTML}{A23B72}
\definecolor{qwenorange}{HTML}{F18F01}
\usepackage{soul}
\usepackage{amsmath}
\usepackage{amssymb}
\usepackage{booktabs}
\usepackage{wrapfig}
\usepackage{hyperref}
\usepackage{subcaption} 
\usepackage[table]{xcolor}

\definecolor{darkblue}{rgb}{0, 0, 0.5}
\hypersetup{colorlinks=true, citecolor=darkblue, linkcolor=darkblue, urlcolor=darkblue}

\title{Can LLM Agents Discover? Evaluating Creativity on ML Engineering Tasks}

\author{Shitanshu Bhushan, Yunxiang Zhang, Lu Wang \\
Computer Science and Engineering \\
University of Michigan \\
Ann Arbor, MI, USA \\
\texttt{\{sbhushan,yunxiang,wangluxy\}@umich.edu}
}

\begin{document}

\ifcolmsubmission
\linenumbers
\fi

\maketitle

\begin{abstract}
Recent AI systems promise autonomous scientific discovery~\citep{mitchener2025kosmosaiscientistautonomous,aiscientist_v2}, claiming to discover algorithms and produce research papers, yet understanding whether they exhibit creativity, the capacity to produce solutions that are both novel and useful, remains an open question. We present a framework for evaluating multi-turn LLM research agents' creativity using ML engineering tasks as a testbed, through three dimensions: \textbf{P-Creativity} (psychological novelty: novel relative to the agent's own prior solutions within a run), \textbf{H-Creativity} (historical novelty: novel relative to the corpus of human solutions), and \textbf{Usefulness} (task performance). Evaluating two agent frameworks, AIDE \citep{aide2025} and AIRA-Dojo \citep{toledo2025airesearchagentsmachine}, on 10 Kaggle-style machine learning tasks from MLE-Bench~\citep{DBLP:conf/iclr/ChanCJASMSLMPMW25}, we develop an LLM-as-a-Judge pipeline and verify its strong correlation with human creativity judgments, providing a reliable automated metric for P-Creativity evaluation at scale. Applying this pipeline to agent trajectories, we find: (1) all agents exhibit declining P-Creativity as they transition from exploration to exploitation; (2) LLMs exhibit greater H-Creativity than medal-winning humans, yet achieve lower performance. Our findings reveal that current agents can explore novel regions of the solution space 
but lack the capacity to convert this novelty into improved task performance.\footnote{Code is available at \url{https://github.com/shitanshubhushan/Creativity-Evaluation-Framework}.}
\end{abstract}

\section{Introduction}
\label{sec:Introduction}
Autonomous LLM agents are increasingly deployed for extended research tasks, with recent systems claiming to discover novel algorithms~\citep{novikov2025alphaevolvecodingagentscientific}, produce conference-ready papers~\citep{aiscientist_v2,zochi2025}, and make breakthrough discoveries~\citep{mitchener2025kosmosaiscientistautonomous}.
Yet on ML engineering benchmarks, even state-of-the-art agents do not consistently match the performance of top humans
~\citep{DBLP:conf/iclr/ChanCJASMSLMPMW25,chen2025mlrbenchevaluatingaiagents,qiang2025mledojointeractiveenvironmentsempowering,wijk2025rebenchevaluatingfrontierai,DBLP:journals/corr/abs-2504-09702}. \textbf{Creativity}, the capacity to produce solutions that are both novel and useful~\citep{runcojaegerstdcreative}, may explain this gap: agents explore regions distant from human solutions, but struggle to make this exploration useful.

Testing this hypothesis requires measuring creativity at scale. Human judgment is the gold standard but does not scale. Prior approaches evaluate single-turn outputs or measure ideation breadth~\citep{audranreiss2025doesgoodairesearch,GUZIK2023100065,sen2025think}, and concurrent work treats novelty as a single scalar against final submissions~\citep{zhang2026innogymbenchmarkinginnovationpotential}, without separating novelty relative to the agent's own experience from novelty relative to the human solution space. This distinction matters: for a solution to be novel to the human corpus, it must first be novel to the agent itself.
We address this gap by evaluating LLM agent creativity in a multi-turn agentic setting on real-world ML competitions.

We ground our evaluation in cognitive science. Following \citet{BODEN1998347}, we distinguish P-creativity (novelty relative to the agent's own prior solutions within a run) from H-creativity (novelty relative to all documented human solutions for the task). Following \citet{Chan02102023}, we decompose usefulness into feasibility (whether solutions execute successfully) and impact (performance improvement on task metrics). Crucially, our framework treats novelty and usefulness as separate dimensions: an agent that produces distant but unhelpful solutions scores high on novelty but low on creativity. 

\begin{figure}[t]
\centering
\begin{subfigure}[b]{0.46\linewidth}
    \centering
    \includegraphics[width=\linewidth]{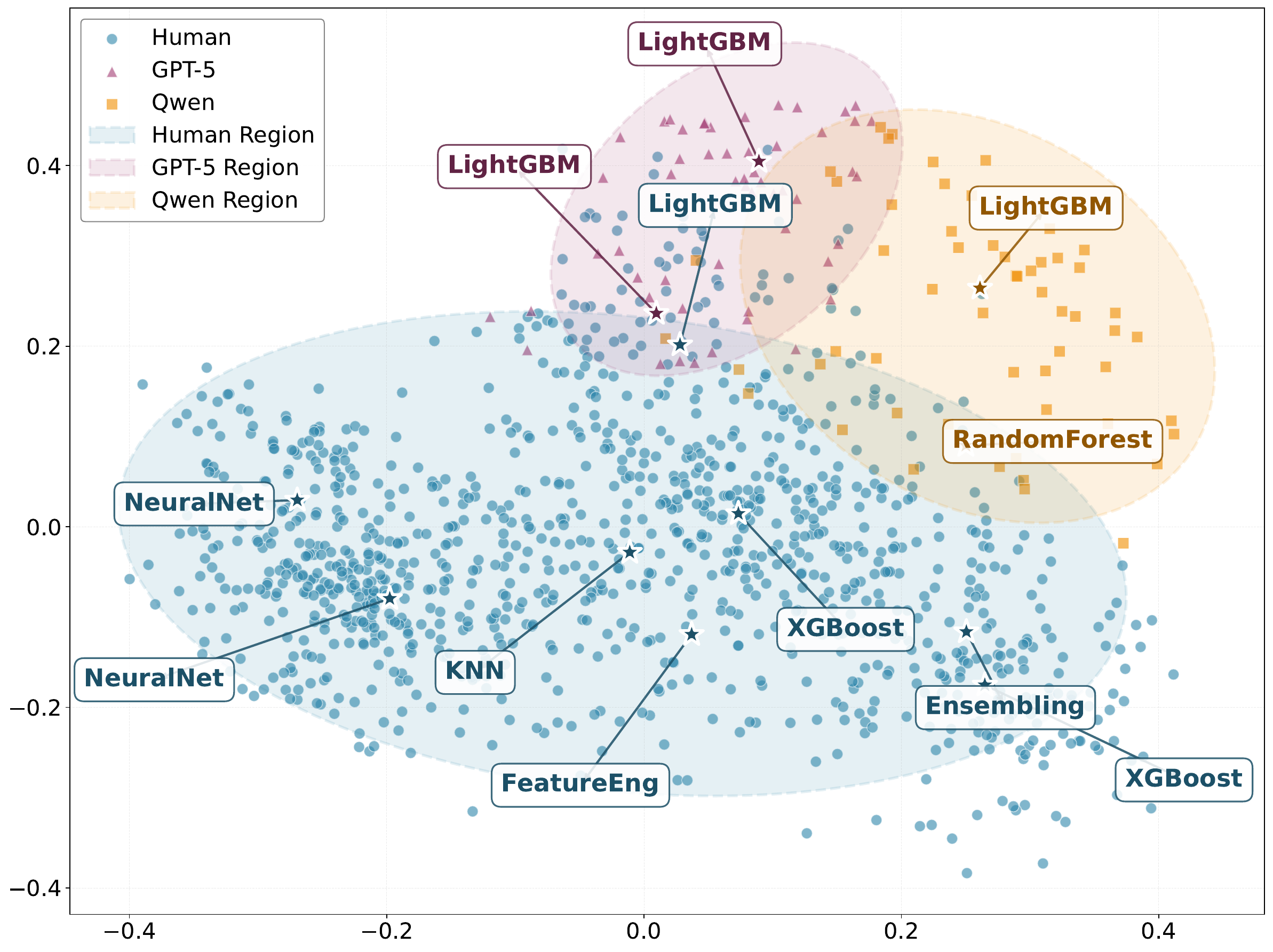}
    \caption{\textbf{LLM agent versus human solution spaces.} PCA visualization of solution embeddings from the Tabular Playground Series Dec 2021 competition in MLE-Bench under the AIDE framework. Points represent solutions from \textcolor{humanblue}{Humans}, \textcolor{gpt5purple}{GPT-5}, and \textcolor{qwenorange}{Qwen3-32B}; stars mark cluster centroids labeled by algorithmic approach.}
    \label{fig:pca_clusters}
\end{subfigure}
\hfill
\begin{subfigure}[b]{0.46\linewidth}
    \centering
    \includegraphics[width=\linewidth]{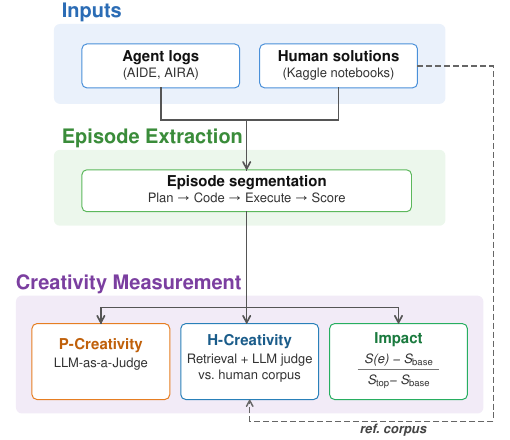}
    \caption{\textbf{Creativity evaluation framework.} Agent logs and human solutions are segmented into episodes, then evaluated along three dimensions: P-creativity (novelty relative to the agent's own trajectory), H-creativity (novelty relative to a human reference corpus), and impact (normalized score improvement).}
    \label{fig:framework_diagram}
\end{subfigure}
\caption{\textbf{Overview of our creativity evaluation approach.} 
(a) Do LLM agents explore novel solution regions or recombine approaches within human solution spaces? Our findings suggest the former. (b) Our framework for measuring P-creativity, H-creativity, and impact.}
\label{fig:introduction}
\end{figure}

We study 10 ML competitions from MLE-Bench~\citep{DBLP:conf/iclr/ChanCJASMSLMPMW25} (see Appendix~\ref{app:taskselection} for selection criteria and details). ML competitions uniquely satisfy three conditions required for meaningful creativity measurement in a multi-turn agentic setting: quantifiable usefulness metrics, rich human baselines for H-creativity comparison (877 to 3,747 public notebooks per competition), and a solution space where genuine novelty is possible. Within this setting, we investigate three research questions:

\begin{itemize}
    \item{\textbf{RQ1:}} Can we reliably measure P-creativity using automated metrics?
    \item{\textbf{RQ2:}} How do P-creativity and impact evolve across iterations for agents versus humans?
    \item{\textbf{RQ3:}} How novel are agent solutions compared to human approaches (H-creativity)?
\end{itemize}

Our key findings reveal: (1) LLM-as-a-judge achieves strong correlation with human P-creativity judgments, providing a reliable automated metric. (2) Across all agents, human and LLM alike, performance improves while P-creativity declines over iterations, with no reliable association between P-creativity and performance.
(3) LLM agents exhibit greater H-creativity than medal-winning humans, yet fail to translate this into performance gains.

Our contributions include: (1) a framework operationalizing creativity through three measurable dimensions; (2) validated automated metrics for P-creativity evaluation at scale; (3) empirical evidence across two agent frameworks and two model families that agents explore beyond human solution spaces yet achieve lower performance, indicating that novelty without usefulness is insufficient for autonomous discovery. 
For practitioners, our validated P-creativity metric is directly actionable: it can serve as an optimization signal via novelty bonuses or adaptive exploration schedules (see Appendix~\ref{app:future_directions}).

\section{Related Work}
\label{sec:relatedwork}

\paragraph{AI for Science}
There is a large and growing body of work in AI for Science, with recent systems generating research ideas, running experiments, and producing papers end-to-end: the AI Scientist and its successor~\citep{lu2024aiscientistfullyautomated, aiscientist_v2}, AlphaEvolve~\citep{novikov2025alphaevolvecodingagentscientific}, Kosmos~\citep{mitchener2025kosmosaiscientistautonomous}, Google's AI co-scientist~\citep{Gottweis_2026}, Zochi~\citep{zochi2025}, DeepScientist~\citep{weng2025deepscientistadvancingfrontierpushingscientific}, EvoScientist~\citep{lyu2026evoscientistmultiagentevolvingai}, AutoResearchClaw~\citep{liu2026autoresearchclawselfreinforcingautonomousresearch}, and others. Recently, OpenAI reported that an internal version of its model made substantial progress on ten open problems in mathematics and theoretical computer science~\citep{openai2026tenadvances}. Yet audits show idea novelty and execution effectiveness decouple once ideas are actually tested~\citep{si2025ideationexecutiongapexecutionoutcomes},
leaving it unclear whether these are isolated successes or evidence of a reliable emergent capability of models.

\paragraph{LLM Creativity Evaluation}
Existing work on LLM creativity evaluation spans psychological assessments~\citep{GUZIK2023100065,zhang2025aideliverscreativeoutput,bellemarepepin2025divergentcreativityhumanslarge}, single-turn automated metrics~\citep{sen2025think,tian-etal-2024-macgyver,zhang2025noveltybenchevaluatinglanguagemodels}, and domain-specific benchmarks like CreativeEval for hardware code~\citep{DBLP:journals/corr/abs-2404-08806}. Recent benchmarks introduce iterative elements: NEOCODER~\citep{lu-etal-2025-benchmarking} applies progressively tighter constraints to code generation, and CreativityPrism~\citep{hou2025creativityprismholisticbenchmarklarge} organizes creativity into quality, novelty, and diversity. However, both evaluate outputs at each state independently rather than tracking how creativity evolves through iterative revision. \citet{audranreiss2025doesgoodairesearch} study ideation breadth in AI research agents on MLE-Bench, and \citet{zhang2026learningideatemachinelearning} train a dedicated ideation module for performance, but neither tracks how novelty and usefulness co-evolve as agents receive feedback and refine solutions. At a meta level, AGC-Bench synthesizes this fragmented landscape via a review of over 3,000 papers, showing across 83 LLMs that creativity forms a coherent factor separable from general reasoning~\citep{beaty2026agcbenchmeasuringartificialgeneral}.

Most directly related are three concurrent works. InnoGym~\citep{zhang2026innogymbenchmarkinginnovationpotential} evaluates agent innovation through performance gain and novelty, but only against a small, curated set of human solutions (1--7 per task), never the agent's own prior attempts. InnovatorBench~\citep{wu2025innovatorbenchevaluatingagentsability} evaluates end-to-end LLM research tasks like loss and reward design, reporting only task performance, with no creativity metric. CreativeBench~\citep{wang2026creativebenchbenchmarkingenhancingmachine} also draws on Boden's framework, but scores a single generation per problem against one reference solution, with no trajectory and no human corpus. More broadly, current work tends to measure novelty and usefulness in isolation, typically as a single generation rather than how creativity evolves across an agent's trajectory, and even then using automated metrics that favor stylistic novelty over conceptual novelty and shift under minor prompt variation~\citep{lu2026rethinkingcreativityevaluationcritical}. Our framework addresses these gaps: grounded in creativity psychology~\citep{BODEN1998347}, we decompose novelty into P-creativity and H-creativity, validate our metrics against human annotations, and track their evolution across a far broader human reference set (877 to 3,747 solutions per task).

\section{Methodology}
\label{sec:Methodology}

We first define our unit of analysis (\S\ref{subsec:episodedef}), then establish the theoretical basis for our creativity framework
(\S\ref{subsec:creativityframework}), and finally detail how novelty is
operationalized and validated for both P-creativity and H-creativity (\S\ref{subsec:pcreativitymetrics}).

\subsection{Unit of Analysis: Episodes}
\label{subsec:episodedef}
Research agents cycle through ideation, generation, execution, debugging, and revision. Evaluating creativity per action is not meaningful; we need a unit that captures a complete problem-solving attempt. We segment trajectories into \textbf{episodes}: an episode begins when the agent proposes a new plan and ends when code executes successfully with a valid submission score. Intermediate failures are part of the same episode. For each episode, we extract: (1) final executed code, (2) agent's natural language plan, (3) performance score, and (4) timestamp. See Appendix~\ref{app:episodes} for details.

\subsection{Creativity Framework}
\label{subsec:creativityframework}

\paragraph{Defining creativity.}
The standard definition of creativity in psychology requires two essential criteria: \textbf{originality} and \textbf{usefulness}~\citep{runcojaegerstdcreative}. Originality refers to the novelty of an idea relative to existing knowledge, while usefulness indicates whether the idea is helpful to some group of individuals. This dual-criterion framework is fundamental: \textbf{novel but impractical ideas are not creative, nor are useful but conventional solutions.} For autonomous research agents, this translates to a requirement that they both explore beyond standard approaches and produce solutions that demonstrably improve task performance.

\paragraph{P-creativity and H-creativity.}
\citet{BODEN1998347} distinguishes between two forms of novelty: \textbf{P-creativity} (psychological creativity) and \textbf{H-creativity} (historical creativity). P-creativity measures the novelty of an idea relative to the system's own prior knowledge and experience. H-creativity evaluates novelty relative to all documented human solutions for the task. In our setting, this corpus is the set of approaches the Kaggle community explored for each task.

This distinction is critical: H-creativity is a special case of P-creativity, as ideas novel to the community must first be novel to the agent itself. P-creativity is the fundamental capacity for research agents, indicating whether an agent explores beyond its prior experience rather than reproducing learned patterns. H-creativity emerges only when this exploration reaches territory the community has not documented.

\paragraph{Impact and feasibility.}
Following \citet{Chan02102023}, we decompose usefulness into two components: Feasibility and Impact. 
\textbf{Feasibility} measures whether an idea can be successfully implemented. For code-generating agents, this encompasses: Does the code execute without errors? Does it satisfy resource constraints? Does it conform to problem specifications? 
\textbf{Impact} quantifies the degree to which a successfully implemented idea improves performance on the target metric.
\begin{equation}
\text{Impact}(e) = \frac{S(e) - S_{\text{baseline}}}{S_{\text{top-1}} - S_{\text{baseline}}}
\end{equation}
where $S(e)$ is the episode's score, $S_{\text{baseline}}$ is the sample submission score, and $S_{\text{top-1}}$ is the top human score on the competition leaderboard.\footnote{We invert for lower-is-better metrics.}
Feasibility is captured by our episode definition (see \S\ref{subsec:episodedef}): only episodes that execute successfully and produce a valid score enter analysis. Across our two model families on the AIDE agent framework, we observe attempt success rates of 45.0\% for GPT-5~\citep{singh2025openai} and 26.2\% for Qwen3-32B~\citep{qwen3technicalreport}.

\subsection{Measuring Novelty} \label{subsec:pcreativitymetrics}
Measuring novelty is an open problem with many competing approaches, each with distinct tradeoffs in cost, scalability, and sensitivity to variations. Human annotation remains the most reliable signal but does not scale beyond small samples. We evaluate three families of automated metrics against human P-creativity annotations to determine which best approximates how humans assess algorithmic novelty, then adopt the validated metrics for our analysis.

\paragraph{Human annotations as ground truth.}
Three trained annotators independently labeled 300 episodes using a 5-point ordinal rubric grounded in Boden's creativity framework: 0 (Routine) through 4
(Transformational). The task is comparative: does this episode represent a meaningfully different approach from prior episodes in the same run? This
scoping reduces the need for deep subfield expertise, since annotators assess relative change rather than absolute novelty against the broader ML
literature. Each rubric level is anchored with ML competition examples (e.g., adjusting a learning rate is Routine; switching from single-model
training to cross-validated ensembling is Exploratory; see Appendix~\ref{app:rubric} for the full rubric). Final labels were determined by majority vote, with ties resolved by the authors (Krippendorff's $\alpha$ = 0.724, ordinal).

\paragraph{Automated metrics.}
We evaluate three families of automated metrics against these human annotations, reporting Spearman correlations for all variants in Table~\ref{tab:q1_correlation} (see \S\ref{subsec:resultsrq1}).

\textbf{LLM-as-a-Judge.} We prompt
GPT-5 with the same 5-point rubric used by human annotators. For each episode $e_i$, the judge receives the episode's plan, code, and summary
alongside all prior episodes $\{e_0, \ldots, e_{i-1}\}$, and outputs a score with rationale.

\textbf{Semantic Distance.} We encode each episode (plan + code) into a dense vector using Qwen3-Embedding-4B~\citep{qwen3embedding}. We test four variants: \textit{nearest neighbor} (minimum cosine distance to any prior episode), \textit{centroid} (distance to the mean of prior embeddings), \textit{mean} (average distance across all prior episodes), and \textit{graph-edit} (insertion cost into a similarity graph).

\textbf{Conceptual Novelty.} We extract algorithmic concepts from each episode using GPT-5-nano and measure novelty via set-theoretic operations: \textit{fuzzy membership} computes Gaussian similarity to prior concepts in embedding space, while \textit{set membership} computes the fraction of entirely new concepts not seen in any previous episode.

Detailed formulations for all metrics appear in Appendix~\ref{app:metrics}.

\paragraph{Scaling to H-creativity.}
P-creativity compares an episode against the agent's own prior attempts, making LLM-as-a-Judge feasible. H-creativity compares against the full Kaggle community corpus for each task (described in \S\ref{sec:expsetup}), where directly judging against every human solution is infeasible. Relying on semantic distance alone for this comparison would risk conflating surface-level difference with genuine algorithmic novelty.
We therefore use a two-stage pipeline: semantic distance retrieves the $k$ nearest neighbors from the human corpus, then GPT-5 judges the agent's approach against these neighbors on a 0\textendash4 novelty scale (Appendix~\ref{app:measure-h-creative}). This combines the scalability of embedding-based retrieval with the semantic sensitivity of LLM evaluation, ensuring that high H-creativity scores reflect substantive methodological divergence rather than implementation artifacts. Both human and agent code pass through the same LLM summarization prompt before embedding, so that retrieval reflects algorithmic substance rather than differences in coding style.
\section{Experimental Setup}
\label{sec:expsetup}

\textbf{Tasks.} 
We evaluate on 10 competitions from MLE-Bench~\citep{DBLP:conf/iclr/ChanCJASMSLMPMW25}, which derives tasks from Kaggle competitions. This provides access to human solution trajectories through public notebooks, enabling direct comparison between agent and human creative processes. Our selected tasks span diverse modalities and difficulty levels; see Appendix~\ref{app:taskselection} for selection criteria, our rationale for choosing Kaggle tasks, and a discussion of how this scope affects generalization to open-ended settings.

\textbf{Agent Frameworks.}
We evaluate two frameworks that represent diverse search strategies in solution space. AIDE~\citep{aide2025} uses greedy tree search, where each node represents a complete solution and the framework selects the highest-scoring node to expand at each step. AIRA-Dojo~\citep{toledo2025airesearchagentsmachine} formalizes agents as search policies over code artifacts and supports multiple search strategies, with a redesigned operator set (scoped memory, adaptive prompt) that distinguishes it from AIDE even under greedy selection. We evaluate three AIRA-Dojo variants: Greedy, MCTS, and Evolutionary. For MCTS we run Qwen3-32B across all 10 tasks (8 runs per task), while Greedy and Evolutionary are run on a subset of 3 tasks (4 runs each) to isolate the effect of search strategy on creativity patterns. For AIDE, we evaluate GPT-5 and Qwen3-32B across all 10 tasks (8 runs per model-task combination). All runs were allocated 8-hour budgets with a maximum of 10 episodes per run.

\textbf{Human Comparison Data.} To compare agents against humans, we leverage the unique advantage of 
MLE-Bench's Kaggle origins to access comprehensive human solution data through Meta-Kaggle~\citep{megan_risdal_timo_bozsolik_2022} and Meta-Kaggle Code~\citep{jim_plotts_megan_risdal_2023}. We construct three human comparison sets: 
\begin{itemize}
    \item \textbf{Episode-wise trajectories (RQ2):} Humans with $\geq$8 submissions spanning $\geq$60\% of the competition timeline, yielding 5--12 eligible participants per competition across 7 competitions. These trajectories enable comparison of how P-creativity and performance co-evolve across episodes between agents and humans.
    \item \textbf{H-creativity reference corpus (RQ3):} All public notebooks with valid scores for each competition (877--3,747 per task) across all competitions. This is the corpus against which H-creativity is measured throughout RQ3: a solution is historically novel if it differs from what practitioners attempted on the same task. With hundreds to thousands of participants per task, including many experienced practitioners, this corpus captures the breadth of approaches humans actually explored.
    \item \textbf{Elite post-competition corpus (RQ3):} Medal-winning notebooks submitted after each competition ended, representing solutions developed with full access to leaderboard results and community knowledge. We score these notebooks with the same H-creativity pipeline against the reference corpus above, using them to benchmark human H-creativity against agent H-creativity.
\end{itemize}
To enable fair comparison, both human and agent code pass through the same
LLM summarization and embedding pipeline
(\S\ref{subsec:pcreativitymetrics}). See
Appendix~\ref{app:creativityinthewild} for further details.
\section{Results}
\label{sec:results}

\subsection{Can Automated Metrics Reliably Measure P-Creativity? (RQ1)}
\label{subsec:resultsrq1}

\begin{wraptable}{r}{0.48\textwidth}
    \vspace{-12pt}
    \begin{center}
    \small
    \resizebox{\linewidth}{!}{\setlength{\tabcolsep}{3pt}
\begin{tabular}{lcc}
    \toprule
    \textbf{Metric} & \textbf{Corr.} & \textbf{95\% CI} \\
    \midrule
    \multicolumn{3}{@{}l}{\textbf{LLM-as-a-Judge}} \\
    \quad \textbf{GPT-5} & \textbf{0.732} & \textbf{[0.675, 0.781]} \\
    \quad GLM-5 & 0.703 & [0.641, 0.756] \\
    \quad DeepSeek V3.2 & 0.700 & [0.637, 0.753] \\
    \quad Qwen3-235B & 0.685 & [0.620, 0.741] \\
    \quad GPT-OSS-120B & 0.544 & [0.460, 0.620] \\
    \midrule
    \multicolumn{3}{@{}l}{\textbf{Semantic distance}} \\
    \quad Nearest neighbor & 0.637 & [0.565, 0.700] \\
    \quad Centroid & 0.589 & [0.510, 0.659] \\
    \quad Mean & 0.503 & [0.413, 0.583] \\
    \quad Graph & $-$0.329 & [$-$0.426, $-$0.224] \\
    \midrule
    \multicolumn{3}{@{}l}{\textbf{Conceptual novelty}} \\
    \quad Fuzzy membership & 0.539 & [0.453, 0.614] \\
    \quad Set membership & 0.508 & [0.419, 0.588] \\
    \bottomrule
\end{tabular}}
    \end{center}
    \caption{\textbf{Spearman correlations between automated metrics and human P-creativity annotations.} Higher values indicate stronger agreement. LLM-as-a-Judge with GPT-5 achieves the strongest agreement with human judgment, outperforming embedding-based approaches. All correlations are significant ($p < 0.001$).}
    \label{tab:q1_correlation}
    \vspace{-10pt}
\end{wraptable}

To evaluate the reliability of automated P-creativity metrics, we compare seven approaches against human annotations (\S\ref{subsec:pcreativitymetrics}). Table~\ref{tab:q1_correlation} shows Spearman correlations across 300 episodes.

\textbf{LLM-as-a-Judge with GPT-5 achieves the highest correlation} ($r = 0.732$), outperforming both other judge models and all embedding-based approaches. Among LLM judges, performance scales with model capability: DeepSeek V3.2~\citep{deepseekai2025deepseekv32}, GLM-5~\citep{glm5team2026glm5}, and Qwen3-235B~\citep{qwen3technicalreport} perform competitively, while the smaller GPT-OSS-120B~\citep{openai2025gptoss120bgptoss20bmodel} performs similarly to embedding-based methods. This scaling suggests P-creativity evaluation benefits from reasoning capabilities that improve with model size, as assessing novelty requires parsing algorithmic intent from code and plans and distinguishing meaningful novelty from superficial variation. These results validate LLM-as-a-Judge as a reliable proxy for human judgment.

To test whether this correlation depends on prompt design, we ran a leave-one-out ablation on the GPT-5 judge, removing each prompt component independently (Table~\ref{tab:prompt_ablation}). All variants remain strongly correlated with human annotations, indicating that LLM-as-a-Judge is not brittle or overly sensitive to prompt design choices.

\textbf{Bias analysis.} We additionally test the GPT-5 judge for three potential sources of bias: \textbf{(1) Verbosity bias:} To test whether the judge rewards or penalizes longer episodes, we compute the partial correlation between judge error (GPT-5 score minus human score) and episode length, controlling for true human score and using plan length or code length as the length covariate; both plan length ($\rho=-0.032$, $p=0.585$) and code length ($\rho=-0.074$, $p=0.201$) are near-zero and non-significant.
The judge does not reward or penalize verbosity. \textbf{(2) Position-index bias:} To test whether judge reliability degrades as episode context accumulates, we measure the correlation between episode position and judge error; both magnitude ($\rho=-0.041$, $p=0.482$, whether the size of judge error increases as episode position advances) and direction ($\rho=+0.049$, $p=0.402$, whether the judge leans more toward over- or under-scoring as episode position advances) are near-zero and non-significant. 
\begin{wraptable}{r}{0.34\textwidth}
    \vspace{-12pt}
    \raggedleft
    \small
    \begin{tabular}{lc}
    \toprule
    \textbf{Variant} & \textbf{Spearman $\rho$} \\
    \midrule
    \textbf{Full prompt}   & \textbf{0.732} \\
    w/o examples  & 0.684 \\
    w/o edge rules & 0.699 \\
    w/o procedure & 0.712 \\
    \bottomrule
    \end{tabular}
    \captionof{table}{\textbf{Prompt sensitivity ablation (GPT-5 judge).} Spearman correlation with human annotations after removing each prompt component independently. Correlation remains strong across all variants, indicating the judge is not brittle to prompt design.}
    \label{tab:prompt_ablation}
    \vspace{-10pt}
\end{wraptable}
Increasing context from prior episodes does not degrade judge reliability. 
\textbf{(3) Self-preference bias:} To test whether GPT-5 systematically favors its own outputs, we use DeepSeek V3.2 as a neutral judge
on all AIDE (GPT-5 and Qwen3-32B) episodes and compute $\Delta$ = GPT-5 score minus DeepSeek score; $\Delta$ is significantly larger for Qwen episodes than GPT-5's own ($+0.300$ vs. $+0.129$, $p=0.000038$). GPT-5 is more conservative on its own outputs and shows no sign of self-preference. No systematic biases were observed across any of the three tests.

\textbf{Practical implications.} LLM-as-a-Judge's superior performance comes with substantial cost: each evaluation requires processing the current episode plus all prior episodes through a frontier LLM (Refer to Appendix~\ref{app:future_directions} for more discussions). For large-scale evaluation, we recommend combining both approaches: semantic distance as a fast retrieval stage to narrow the comparison set, followed by LLM-as-a-Judge on the filtered candidates. This two-stage strategy is what
motivates our H-creativity pipeline (\S\ref{subsec:pcreativitymetrics}).

\subsection{How Do Creativity and Performance Evolve Across Iterations? (RQ2)}
\label{subsec:resultsrq2}

\begin{figure*}[t]
    \begin{center}    \includegraphics[width=0.95\textwidth]{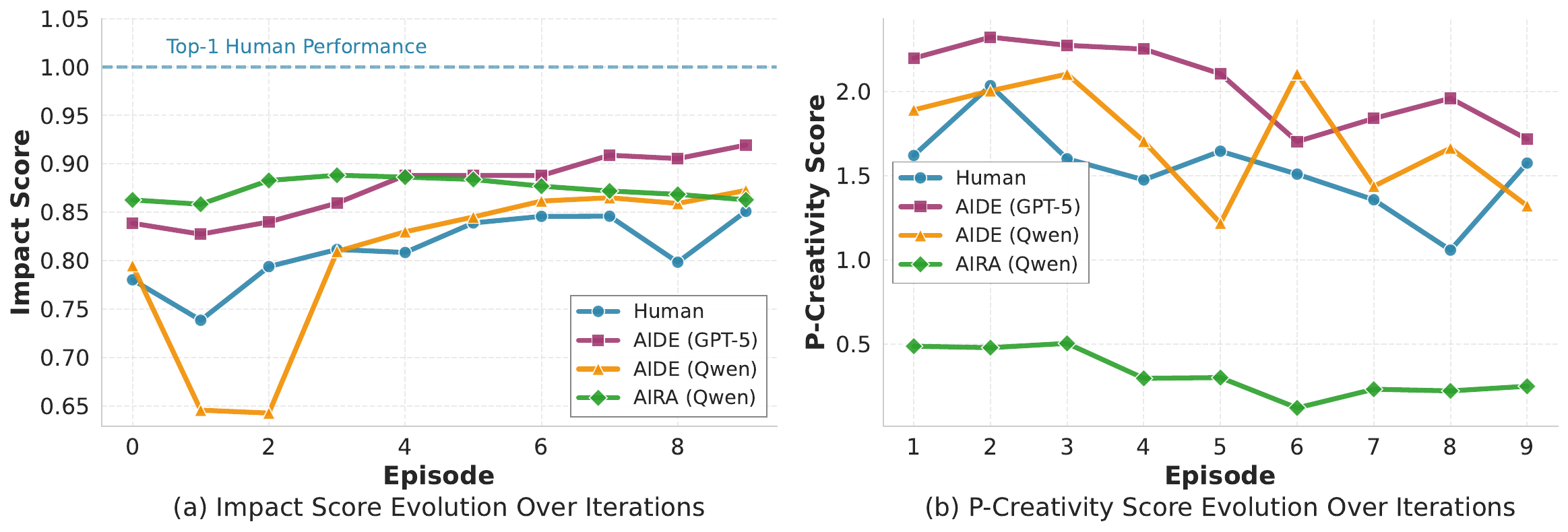}
    \end{center}
    \caption{\textbf{Comparative analyses of impact and P-creativity across episodes.} (a) All agents improve performance, with AIDE (GPT-5) most consistent and AIRA-MCTS (Qwen) starting higher but plateauing. (b) P-creativity declines universally, but AIRA-MCTS (Qwen) operates at persistently lower levels throughout. 
    Note that Plot (b) starts at episode 1, with episode 0 serving as the baseline for P-creativity comparison.}
    \label{fig:rq2_results}
\end{figure*}

Ideally, agents would balance exploration and exploitation: sustaining novelty when a paradigm has headroom, and converging on refinement only when a promising approach warrants it. To test whether iterative refinement achieves this balance,
we analyze trajectories across episodes comparing LLM agents against human problem-solvers\footnote{These humans may not be top-1 performers, but their trajectories represent authentic creative problem-solving that successfully improved over time, providing a meaningful baseline for interpreting agent behavior.} 
from the original Kaggle competitions (Figure~\ref{fig:rq2_results}).

\textbf{Iterative refinement reliably improves impact across all agents.} Figure~\ref{fig:rq2_results}(a) shows steady upward trends across all configurations, confirming that multi-turn agentic search is an effective strategy for performance improvement regardless of model or framework. AIDE (GPT-5) shows the most consistent growth, suggesting that the stronger the reasoning capacity of a model, the more reliable its episode-to-episode gains.

\begin{figure}[t]
    \begin{center}
    \includegraphics[width=0.85\linewidth]{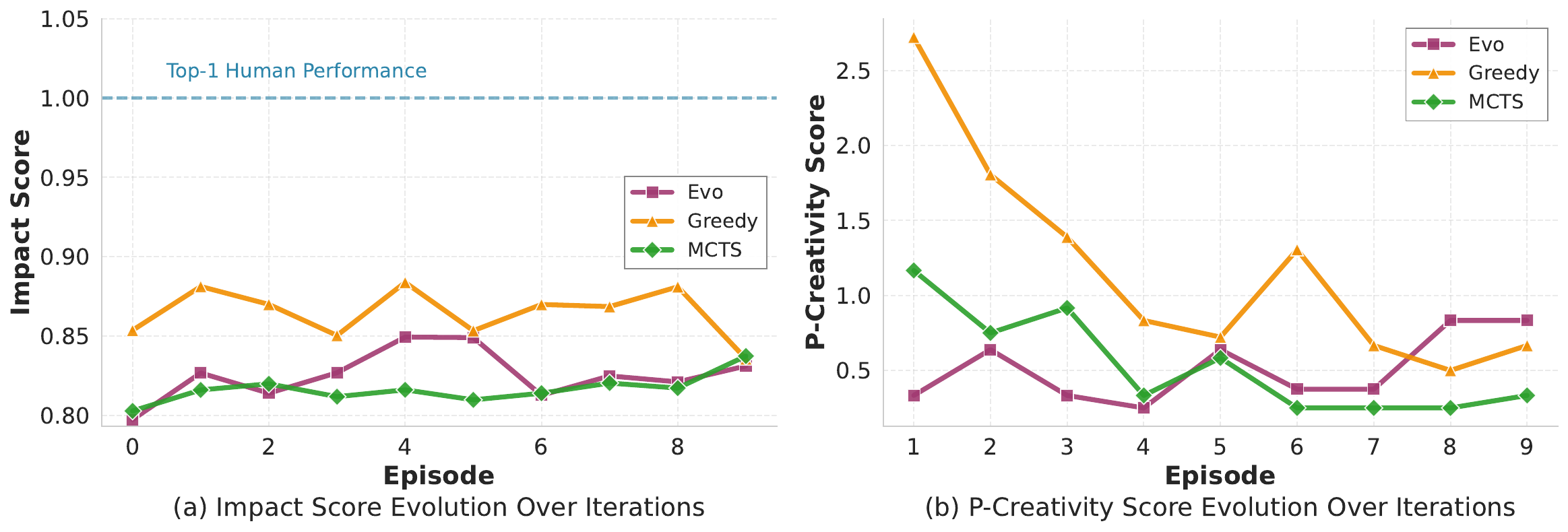}
    \end{center}
    \caption{\textbf{Search strategy comparison within AIRA-Dojo
    (Qwen3-32B, 3 tasks).} Greedy search starts with the highest
    P-creativity but declines steeply. MCTS and evolutionary search strategies 
    maintain lower but more stable P-creativity. Greedy search strategy also achieves the highest
    impact.} 
    \label{fig:rq2_search}
\end{figure}

\textbf{P-creativity declines universally, revealing a structural shift from exploration to exploitation.} Figure~\ref{fig:rq2_results}(b) reveals a fundamental pattern: all agent types exhibit declining P-creativity as episodes progress, indicating a universal shift from exploration to exploitation.
However, the rate and magnitude of decline differ substantially. LLM agents converge more quickly and steeply toward exploitation compared to humans. Notably, AIRA-MCTS operates at persistently low P-creativity from the start, suggesting that the AIRA-Dojo framework prioritizes metric improvement over algorithmic exploration. This is consistent with the search strategy comparison (Figure~\ref{fig:rq2_search}), where all three AIRA-Dojo strategies drop below a P-creativity score of 1 within the first few episodes.

\textbf{P-creativity and performance improvement are decoupled.} 
To test whether exploring novel approaches yields performance gains, we measure
Spearman correlations between episode-to-episode changes in impact and P-creativity ($\Delta\text{Impact} = \text{Impact}(e) - \text{Impact}(e{-}1)$, analogously for $\Delta\text{P-Creativity}$). Correlations are weak across agents (Human: $r= -0.111$*; AIDE GPT-5: $r= -0.021$; AIDE Qwen3-32B: $r= 0.056$; AIRA Qwen3-32B: $r= -0.026$; * indicates $p < 0.05$). This reveals that optimizing for one does not improve the other.

\begin{wrapfigure}{r}{0.48\textwidth}
    \vspace{-12pt}
    \begin{center}
    \includegraphics[width=0.46\textwidth]{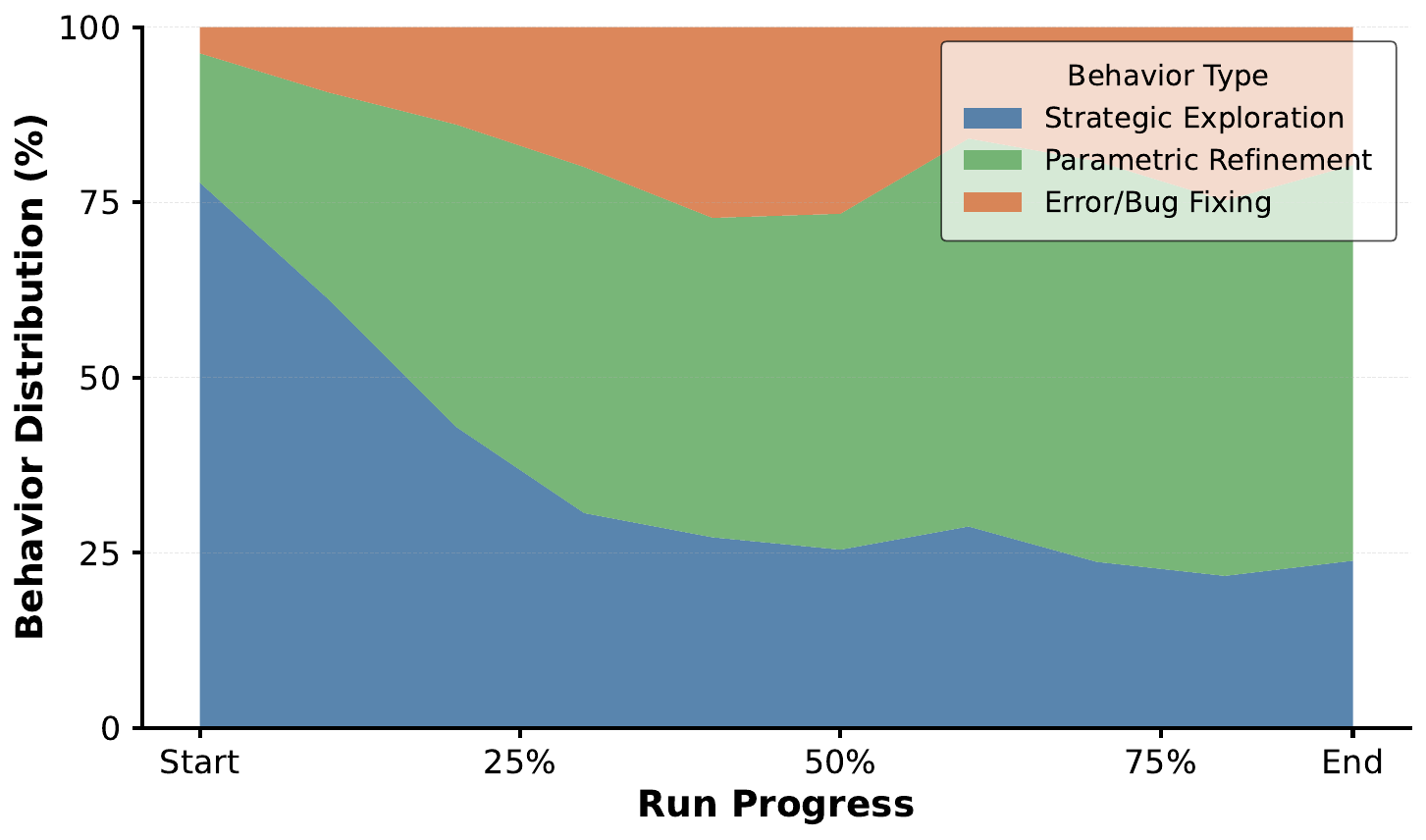}
    \end{center}
    \vspace{-8pt}
    \caption{\textbf{Cognitive behavior distribution across run progression for AIDE (Qwen3-32B).} Strategic exploration dominates early iterations but steadily declines, while parametric refinement grows to dominate later, showing exploration to exploitation trend.}
    \label{fig:cognitive}
    \vspace{-10pt}
\end{wrapfigure}

\textbf{Cognitive behavior analysis confirms the exploration-to-exploitation shift.} To better understand agent thinking, we classified reasoning traces from AIDE (Qwen3-32B) into three behavior types using DeepSeek-V3.2: strategic exploration (considering fundamentally different approaches), parametric refinement (tuning within the current approach), and bug fixing (Figure~\ref{fig:cognitive}; Appendix~\ref{app:cognitive}). Strategic exploration accounts for $\sim$75\% of behavior instances early in runs but drops to $\sim$25\% by run end, while parametric refinement rises from $\sim$15\% to over 50\%. Most strategic thinking occurs in the first quarter; after that, agents commit to refining their initial approach. This aligns with the P-creativity decline: once agents find a working solution, they shift toward refinement. 

\begin{figure}[t]
    \begin{center}
    \includegraphics[width=0.90\linewidth]{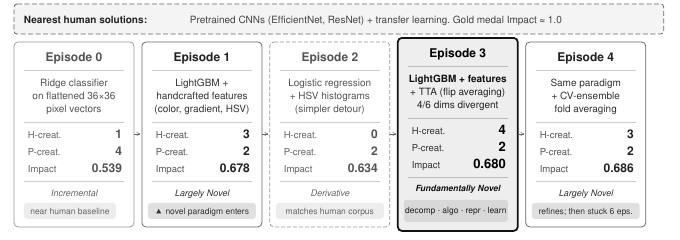}
    \end{center}
    \caption{\textbf{Case study: refinement within a novel but low-ceiling paradigm.} Five consecutive episodes from a single AIDE (GPT-5) run on cassava leaf disease classification. The dashed box shows the nearest human solutions. The agent starts with a conventional approach (Episode 0) but quickly explores a novel region, peaking at H-creativity 4 (Episode 3). Impact initially increases but then stays relatively flat.}
    \label{fig:rq3_casestudy}
\end{figure}

\textbf{Search strategy alone does not determine creativity or impact.} 
Figure~\ref{fig:rq2_search} shows that despite different starting points, all three AIRA-Dojo strategies converge to similar P-creativity and impact levels within a few episodes. This is counter-intuitive: one would expect smarter search to yield meaningfully different exploration patterns. The convergence suggests that search policy is not the binding constraint; the entire agentic scaffolding surrounding it, including how solutions are mutated, how context is passed between nodes, and what the effective action space is, dominates long-run behavior regardless of which policy navigates it. For practitioners, this is an important caution: investing in more sophisticated search strategies may yield diminishing returns without corresponding improvements to the underlying scaffolding.

\subsection{How Novel Are Agent Solutions? (RQ3)}
\label{subsec:resultsrq3}

\begin{wrapfigure}{r}{0.48\textwidth}
    \vspace{-12pt}
    \begin{center}
    \includegraphics[width=0.40\textwidth]{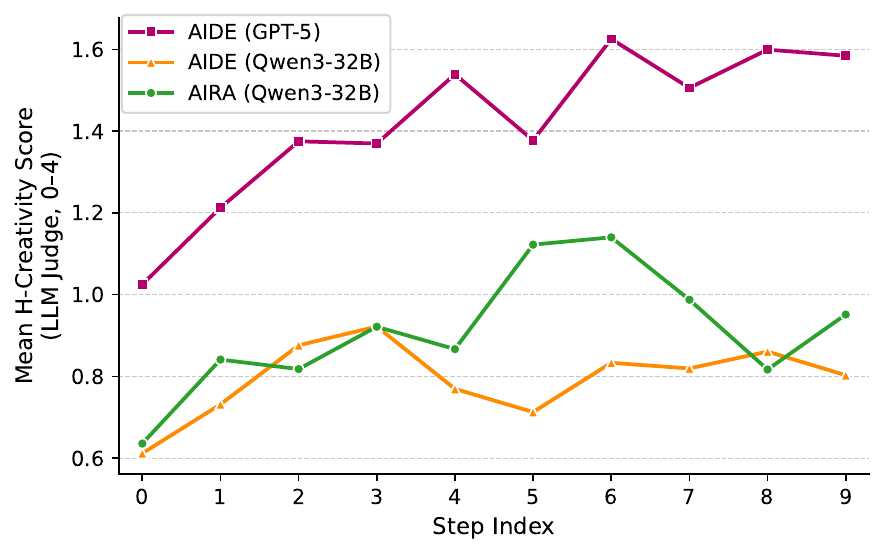}
    \end{center}
    \vspace{-8pt}
    \caption{\textbf{H-creativity evolution across episodes.} GPT-5 diverges further from human solutions as iterations progress. Both Qwen3-32B configurations remain closer to human baselines throughout.}
    \label{fig:RQ3_trajectory}
    \vspace{-10pt}
\end{wrapfigure}

\textbf{GPT-5 explores significantly more novel solution spaces than medal-winning humans, while Qwen3-32B remains comparable.} Among medal-winning human solutions submitted post-competition, H-creativity increases with medal tier (gold: 0.744, silver: 0.524, bronze: 0.293), indicating that novelty and usefulness are positively correlated for humans. GPT-5 with AIDE achieves nearly double the novelty of gold-medal humans (1.423; $p < 10^{-15}$), whereas both Qwen3-32B configurations (AIDE-Greedy and AIRA-MCTS) are statistically indistinguishable from gold-medal humans (AIDE: 0.838, $p = 0.33$; AIRA-Dojo: 0.800, $p = 0.99$). Despite this, only 21.25\% of GPT-5 runs and 10.85\% of AIDE Qwen3-32B runs achieve medal-level performance. This suggests that even when agents explore more novel regions of the solution space, this exploration does not prove more impactful than humans.
The three configurations show distinct trajectories (Figure~\ref{fig:RQ3_trajectory}). GPT-5 exhibits a clear upward trend, diverging further from human approaches as iterations progress, suggesting stronger reasoning capacity drives continued exploration of novel solution regions. AIDE Qwen3-32B remains relatively flat across episodes, while AIRA Qwen3-32B shows higher variance, peaking in the middle episodes before declining.

\textbf{Agent novelty reflects task-specific exploration, not undirected divergence.} A central concern is whether high H-creativity simply reflects being ``different'' rather than genuinely ``novel.'' To calibrate our metric, we construct a cross-task baseline: for each competition, agent episodes generated for a different-modality competition are scored against the target competition's human corpus using the same two-stage pipeline. These are coherent ML solutions to the wrong problem, establishing an upper bound on task-irrelevant divergence. If agent novelty were undirected, in-task scores should approach this ceiling; instead, cross-task episodes score 2.39 on average (higher H-creativity is more novel), compared to 1.42 for AIDE(GPT-5), 0.84 for AIDE(Qwen3-32B), and 0.80 for AIRA(Qwen3-32B) on their actual tasks. This ordering confirms that agent novelty reflects genuine algorithmic positioning rather than undirected divergence.

\begin{wrapfigure}{r}{0.48\textwidth}
    \vspace{-12pt}
    \begin{center}
    \includegraphics[width=0.46\textwidth]{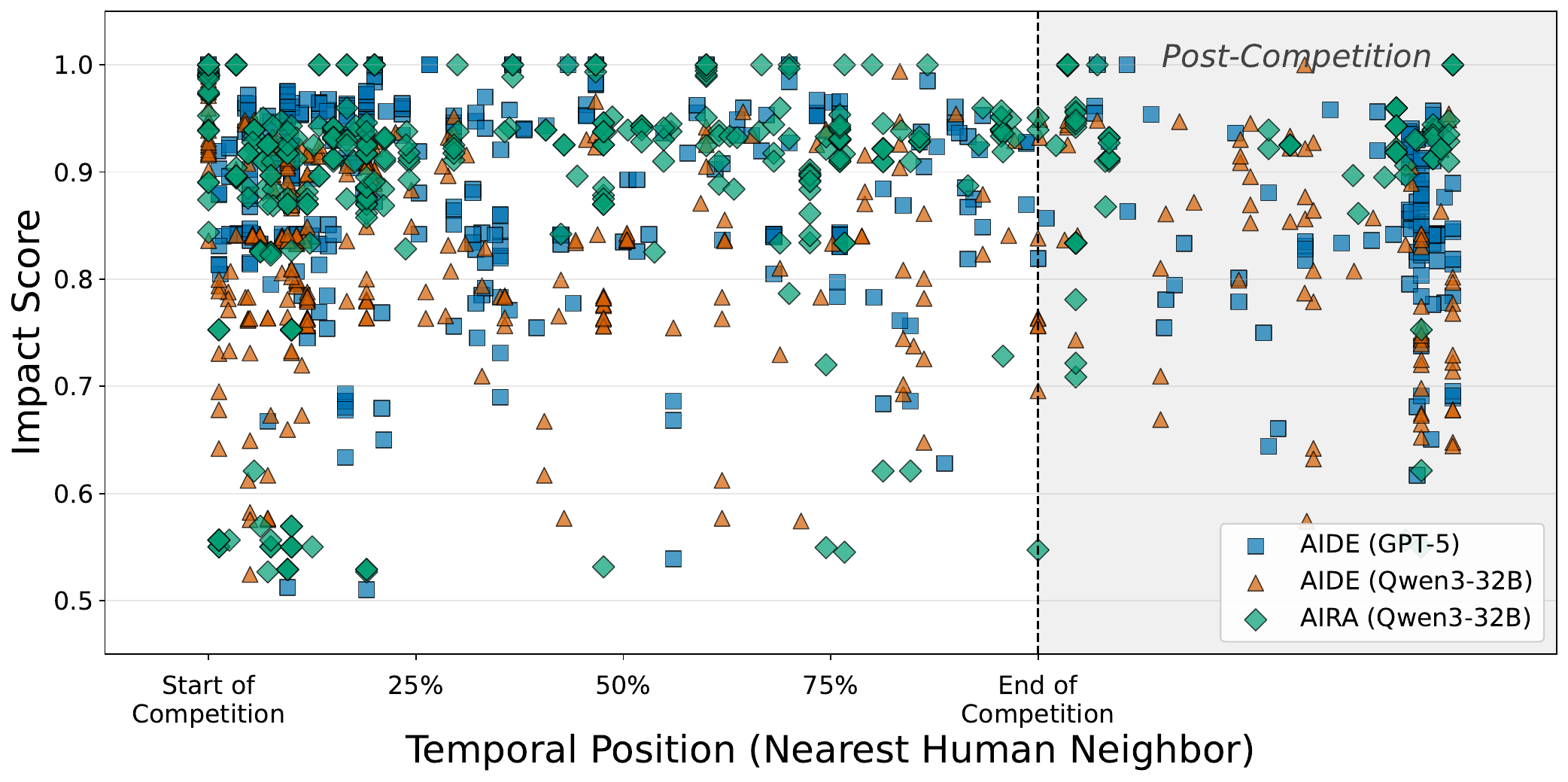}
    \end{center}
    \vspace{-8pt}
    \caption{\textbf{Temporal position of each agent episode's nearest human neighbor vs.\ impact score.} Temporal position reflects when the nearest human neighbor was submitted during the competition timeline. Agent neighbors span the full timeline.}
    \label{fig:RQ3_temporal_bins}
    \vspace{-10pt}
\end{wrapfigure}

\textbf{Agent nearest neighbors are distributed across the full competition timeline.}
A potential concern is that agent novelty reflects regression to early approaches humans later abandoned, rather than forward-looking exploration. To test this, we locate each agent episode's nearest human neighbor by cosine similarity over embeddings of approach summaries on the competition timeline (Figure~\ref{fig:RQ3_temporal_bins}). If agents were regressing to abandoned ideas, nearest neighbors would cluster heavily in the early competition period. Instead, all three configurations show partial concentration in the first quartile (32--44\% of episodes), but this accounts for less than half: 21--23\% of nearest neighbors fall in the post-competition region. GPT-5 shows the smallest early-stage concentration (32.0\%) and the largest late/post-competition share (38.7\%), suggesting that stronger reasoning may enable convergence towards more mature human solutions.

\textbf{Agents refine novel approaches rather than abandoning them, but neither strategy resolves the novelty-impact gap.} To understand the disconnect between novelty and impact, we analyze 92 episodes in both the top 20th percentile of H-creativity and the bottom 20th percentile of impact (Appendix~\ref{app:highHanalysis}). After reaching these novel states, 87.0\% of episodes led to refinement behavior rather than abandoning the current approach. Refinement neither sustained novelty nor improved impact: 53.8\% of refined episodes showed decreased H-creativity in the subsequent episode, with no measurable effect on impact (median $\Delta$Impact$= 0.000$). This pattern shows that agents lack the ability to convert novel solution regions into performance gains regardless of whether they refine or abandon their approach. 

Figure~\ref{fig:rq3_casestudy} walks through a single run that captures the dynamics we report in RQ2 and RQ3. The agent explores productively in early episodes, discovering a genuinely novel paradigm by Episode 3. It then commits to refining this approach through Episode 9 without meaningful improvement: P-creativity drops to 0 and the same accuracy repeats for six consecutive episodes. The agent lacks a signal that its current paradigm has a ceiling, and its search strategy provides no mechanism to return to exploration once refinement stalls. 
\section{Conclusion}
\label{sec:conclusion}
We presented a framework for evaluating creativity in LLM research agents through P-creativity, H-creativity, feasibility, and impact. Evaluating three families of automated metrics against human annotations, we find that LLM-as-a-Judge correlates most strongly with human P-creativity judgments (r = 0.732), outperforming embedding-based and concept-based approaches. Applying this metric across two agent frameworks and two model families, we find that P-creativity declines universally as agents shift from exploration to exploitation, yet this confers no systematic advantage: changes in P-creativity and impact are decoupled, and the decline persists regardless of search strategy. GPT-5 explores solution regions more novel than gold-medal humans, yet only 21.25\% of its runs reach medal-level performance. Together, these findings show that \textbf{neither pure exploitation nor pure exploration alone produces breakthrough discovery}: genuine autonomous discovery requires \textbf{creativity}, the capacity to generate solutions that are both novel and useful. Our P-creativity metric offers a concrete signal that future frameworks can optimize toward.

\section{Acknowledgments}
\label{sec:Acknowledgment}
We thank members of the LAUNCH Lab and COLM reviewers for helpful feedback. This research was supported in part by the National Science Foundation through grant 2046016 and through computational resources and services provided by Advanced Research Computing at the University of Michigan, Ann Arbor. 
This work also used Bridges-2 at Pittsburgh Supercomputing Center through allocation CIS250174 from the Advanced Cyberinfrastructure Coordination Ecosystem: Services \& Support (ACCESS) program, which is supported by National Science Foundation grants \#2138259, \#2138286, \#2138307, \#2137603, and \#2138296.

\bibliography{colm2026_conference}
\bibliographystyle{colm2026_conference}

\appendix
\section{Future Directions and Limitations}
\label{app:future_directions}

\paragraph{P-creativity as an optimization target.} P-creativity decline is structural: agents converge toward refinement regardless of framework or model. Since P-creativity can be reliably measured (RQ1), it is the most actionable optimization target. Concretely, this suggests novelty bonuses penalizing solutions too similar to prior episodes, adaptive exploration schedules that increase diversity when P-creativity drops below a threshold, and dual objectives jointly targeting usefulness and diversity.

\paragraph{Scaling creativity evaluation.} Our evaluation handles trajectories of up to 9 episodes, keeping full history within most LLM context windows. For longer trajectories, this approach becomes computationally infeasible. For such scenarios, context management strategies such as summarizing early episodes could preserve the essential comparison set. More fundamentally, an agent-as-a-judge~\citep{zhuge2024agent} approach, where the evaluating agent maintains a compressed state representation of prior episodes would enable P-creativity measurement over arbitrarily long trajectories. In preliminary tests extending AIRA-MCTS trajectories to 30 episodes, P-creativity remained flat rather than recovering with additional compute, though confirming this pattern across other agent-model configurations remains open.

\paragraph{Extending to open-ended tasks.} Our framework depends on two properties of ML engineering tasks: a quantitative performance metric that operationalizes usefulness, and a bounded human solution corpus that grounds H-creativity. In open-ended research settings, neither is readily available. Extending creativity evaluation to such settings will require surrogate usefulness signals and richer reference corpora, and remains an important open problem.

\paragraph{Data Leakage.} Since MLE-Bench tasks derive from public Kaggle competitions, LLMs may have encountered solution notebooks during training. However, our H-creativity analysis provides evidence against pure strategy memorization. While complete isolation from training data cannot be verified, future work could further validate these patterns on competitions released after model training cutoffs.

\paragraph{H-creativity corpus completeness.} H-creativity in our framework is defined relative to what the Kaggle community attempted on a specific task, not relative to the ML literature at large. This is a deliberate operationalization of Boden's original definition: since novelty relative to all documented human knowledge is unmeasurable in practice, the public notebook corpus is the closest tractable proxy, representing the most comprehensive record of approaches humans actually attempted on exactly this problem. Within this scope, our corpus of 877 to 3,747 notebooks per task is two to three orders of magnitude larger than comparable work, and agents and humans are scored against the same reference set, so residual incompleteness affects both conditions equally rather than biasing the comparison between them. Future work could extend the reference corpus beyond Kaggle notebooks to approximate the fuller space of documented human knowledge that Boden's original definition targets.

\section{Detailed Methodology}
\label{app:methodology}

\subsection{P-Creativity Annotation Rubric}
\label{app:rubric}
To establish ground truth, three trained annotators (two Master's students and one senior undergraduate at a large research university, all with experience in ML and familiarity with Kaggle-style competitions) independently evaluated 300 episodes using a 5-point ordinal rubric (0-4) based on Boden's creativity framework~\citep{BODEN1998347}:

\begin{itemize}
    \item \textbf{0 (Routine)}: Repeating prior approaches with minimal or no variation
    \item \textbf{1 (Combinational)}: Recombining existing components in new configurations
    \item \textbf{2 (Exploratory)}: Introducing novel elements within the established conceptual framework
    \item \textbf{3 (C+E Hybrid)}: Simultaneously recombining known elements and introducing genuinely new ones
    \item \textbf{4 (Transformational)}: Fundamentally reframing the problem representation, objective, or search space
\end{itemize}

Annotators were instructed to evaluate each episode's novelty \textit{relative to all previous episodes in the same run}, focusing on algorithmic and methodological substance rather than code style or formatting. Prior to independent annotation, annotators completed a training session on a held-out set of episodes to establish shared understanding of the rubric.

Final labels were determined by majority vote across the three annotators. Cases with no majority vote were annotated by the authors and used as ground truth. The annotators demonstrated substantial inter-annotator agreement with Krippendorff's $\alpha$ (ordinal) = 0.724, indicating reliable human judgment despite creativity's inherent subjectivity.

\subsection{Detailed Metric Formulations} 
\label{app:metrics}

We evaluated seven automated metrics designed to capture episodic novelty.

\subsubsection{LLM-as-a-Judge}
Based on many recent works~\citep{bavaresco2025llmsinsteadhumanjudges,zheng2023judging} which use LLMs to imitate human judgment, we prompt GPT-5 (\texttt{gpt-5-2025-08-07}) with the P-creativity annotation rubric and task it with scoring each episode $e_i$ on a 0-4 scale by comparing its plan and code against all previous episodes $\{e_0, e_1, \ldots, e_{i-1}\}$. The prompt includes:

\begin{enumerate}
    \item The creativity rubric with detailed descriptions for each score level (0: Routine, 1: Combinational, 2: Exploratory, 3: C+E Hybrid, 4: Transformational)
    \item The current episode's plan, code, and LLM summary (if available)
    \item All previous episodes' plans, codes, and LLM summaries for context
    \item Instructions to output a JSON response with score (0-4), label, Boden modes, and rationale
\end{enumerate}

Each episode $e_i$ contains the code that generated the output, the summary output of its approach by the agent, and an LLM generated summary of the approach in that episode. Details of the Prompts are given in Appendix~\ref{app:promptllmjudge}.

\textbf{Context Window Management:} When the concatenated context of all previous episodes exceeds a token limit of 128,000, we progressively drop code sections from the earliest episodes while retaining their plans and summaries, ensuring the judge still has access to algorithmic approaches.

\subsubsection{Semantic Distance}
Another common approach to measuring novelty is to use the semantic distance from previous approaches. We measure four types of semantic distance to see which best correlates with human annotation.

All semantic distance metrics use \textbf{Qwen/Qwen3-Embedding-4B}~\citep{qwen3embedding} to encode episodes into dense vector representations. This model was selected for its strong performance on code-related tasks in the MTEB leaderboard~\citep{enevoldsen2025mmtebmassivemultilingualtext}. For each episode $e_i$, we extract text containing both the agent's natural language plan and the final executed code, then generate an embedding $\mathbf{v}_i \in \mathbb{R}^d$. Similarity between episodes is measured using cosine similarity:

\begin{equation}
\text{sim}(\mathbf{v}_i, \mathbf{v}_j) = \frac{\mathbf{v}_i \cdot \mathbf{v}_j}{\|\mathbf{v}_i\| \|\mathbf{v}_j\|}
\end{equation}

The novelty score is then $1 - \text{sim}(\mathbf{v}_i, \mathbf{v}_j)$ for distance-based metrics. 

\paragraph{Semantic Distance - Nearest Neighbor (NN)}
For episode $e_i$, compute the minimum cosine distance to any previous episode:

\begin{equation}
\text{P-Creativity}_{\text{NN}}(e_i) = 1 - \max_{j < i} \text{sim}(\mathbf{v}_i, \mathbf{v}_j)
\end{equation}

This captures the intuition: ``How different is this episode from the most similar prior attempt?''

\paragraph{Semantic Distance - Centroid}

Compute the centroid of all previous episode embeddings and measure distance to it:

\begin{equation}
\mathbf{c}_i = \frac{1}{i} \sum_{j=0}^{i-1} \mathbf{v}_j
\end{equation}

\begin{equation}
\text{P-Creativity}_{\text{centroid}}(e_i) = 1 - \text{sim}(\mathbf{v}_i, \mathbf{c}_i)
\end{equation}

\paragraph{Semantic Distance - Mean}
Average cosine distance across all previous episodes:

\begin{equation}
\text{P-Creativity}_{\text{mean}}(e_i) = \frac{1}{i} \sum_{j=0}^{i-1} \left(1 - \text{sim}(\mathbf{v}_i, \mathbf{v}_j)\right)
\end{equation}

\paragraph{Semantic Distance - Graph Edit Cost}
Construct a similarity graph $G = (V, E)$ where vertices $V$ are previous episodes $\{e_0, \ldots, e_{i-1}\}$ and edges exist when similarity exceeds threshold $\tau = 0.4$. The graph edit cost of inserting $e_i$ includes:

\begin{enumerate}
    \item Adding vertex $e_i$ (cost = 1)
    \item Adding edges to all sufficiently similar previous episodes
\end{enumerate}

Higher cost indicates more connections to existing episodes, suggesting lower novelty (the episode is more similar to the existing set). Graph edit cost as operationalized captures the wrong notion of novelty as highly connected episodes have high cost but may represent creative combinations.

\subsubsection{Conceptual Metrics}

Another common approach in creativity measurement is to extract high-level concepts or algorithmic components from each episode and measure novelty through set-theoretic operations~\citep{lu-etal-2025-benchmarking}. The intuition is that creativity can be captured by tracking when new concepts appear.

\textbf{Concept Extraction.} For each episode, we extract algorithmic concepts using GPT-5-nano (gpt-5-nano-2025-08-07) with a structured prompt (Detailed in Appendix \ref{app:conceptextraction}). The model is instructed to identify high-level ideas being implemented in the code and plan, returning 3-10 comma-separated concepts as concise nouns or gerunds (e.g., "XGBoost", "feature engineering", "ensemble", "cross-validation"). This approach provides consistent, semantically meaningful concept extraction across episodes. Each episode $e_i$ is represented by a concept set $C_i$ derived from the extracted comma-separated list.

\paragraph{Fuzzy Set Membership.} We model each concept as a point in embedding space using Qwen3-Embedding-4B and compute fuzzy membership based on semantic similarity to previous concepts. For each concept $c$ in the current episode $e_i$ with embedding $\mathbf{v}_c$, we compute its distance to the nearest previous concept:

$$d_c = \min_{j < i, c' \in C_j} \left(1 - \text{sim}(\mathbf{v}_c, \mathbf{v}_{c'})\right)$$

where $\text{sim}$ is cosine similarity between normalized embeddings. The fuzzy membership is then:

$$\mu_c = \exp\left(-\frac{d_c^2}{\sigma^2}\right)$$

where $\sigma$ is a bandwidth parameter (default: 0.3, or auto-estimated as the median nearest-neighbor distance among previous concepts if sufficient history exists). The per-concept novelty is $1 - \mu_c$, and the episode's P-creativity score is the average novelty across all its concepts:

$$\text{P-Creativity}_{\text{fuzzy}}(e_i) = \frac{1}{|C_i|} \sum_{c \in C_i} (1 - \mu_c)$$

This metric rewards episodes introducing concepts semantically distant from all previous concepts, capturing creative exploration in the conceptual space.

\paragraph{Set Membership.} For a stricter notion of novelty, we compute the fraction of concepts in the current episode that have never appeared in any previous episode. Concepts are first normalized (lowercased, with underscores and hyphens replaced by spaces) to handle minor variations. Let $\hat{C}_i = \bigcup_{j < i} C_j$ denote all concepts seen in episodes before $e_i$:

$$\text{P-Creativity}_{\text{set}}(e_i) = \frac{|C_i \setminus \hat{C}_i|}{|C_i|}$$

This metric rewards episodes that introduce entirely new algorithmic components or strategies not mentioned in any previous episode. 

All of these concept-level metrics are highly sensitive to how concepts are extracted. When there is no standard concept vocabulary, the random variability introduced by LLM extraction can distort the novelty measurement.

\subsubsection{Other metrics considered but not measured}
Apart from these seven metrics, we also explored a few other common approaches and ultimately decided not to use them for various reasons.

\paragraph{Semantic Entropy.}
This is a very common approach to measure novelty and was used by \cite{sen2025think} to measure P-creativity. The basic idea here is that for each input to the LLM, we sample multiple times, cluster them and then find entropy of the cluster probabilities. The issue with this approach is that it is very sample inefficient and would increase the computation needed by a lot in our setting so it is infeasible.

\paragraph{Surprisal.}
Surprisal measures novelty as the negative log probability of an observation given the agent's current memory state. However, surprisal depends on what the agent currently remembers, not what it has actually tried. If an agent explores approach B at step 2, drops it from context due to memory constraints, then revisits the same region at step 6, surprisal would be high despite low P-creativity. This makes surprisal unsuitable for evaluating creativity across full trajectories where context management can decouple perceived novelty from actual novelty.

\subsection{Episode Extraction Details}
\label{app:episodes}

Episodes are extracted from agent logs, which record the complete sequence of actions including planning, code generation, debugging attempts, and execution results. 

\textbf{Episode boundary detection:} An episode begins when the agent generates a new plan and ends when code executes successfully and produces a valid submission score. Failed execution attempts (syntax errors, runtime exceptions, timeouts) are considered part of the episode's trajectory rather than separate episodes.

\textbf{Extracted features:} For each episode, we extract:
\begin{itemize}
    \item \textbf{Code:} The final executed code that produced the valid score
    \item \textbf{Plan:} The agent's natural language reasoning, plan, or approach description
    \item \textbf{Score:} The performance metric value (accuracy, F1, Jaccard, etc.)
    \item \textbf{Timestamp:} When the episode completed
    \item \textbf{Summary:} An LLM-generated summary of the approach
\end{itemize}

\section{Task-Selection and Competition Details}
\label{app:taskselection}
\begin{table*}[t]
    \centering
    \small
    \setlength{\tabcolsep}{4pt} 
    \caption{Selected competitions from MLE-Bench spanning diverse modalities and difficulty levels. Highlighted rows indicate competitions used for episode-wise trajectory analysis (RQ2), where eligible participants have 8+ kernel submissions spanning 60\%+ of competition timeline. All 10 competitions are used for H-creativity evaluation (RQ3).}
    \label{tab:competitions}
    \begin{tabular}{p{0.23\textwidth}cccccc}
\toprule
\textbf{Competition Name} & \textbf{Public} & \textbf{Eligible} & \textbf{Dataset} & \textbf{Modality} & \textbf{MLE-Bench} & \textbf{Competition} \\
 & \textbf{Notebooks} & \textbf{People} & \textbf{Size} &  & \textbf{Level} & \textbf{Type} \\
\midrule
\rowcolor{gray!20} cassava-leaf-disease-classification & 3747 & 12 & 6.19G & Image & Medium & Research \\\cmidrule{1-7}
\rowcolor{gray!20} learning-agency-lab-essay-scoring-2 & 1723 & 5 & 36.2M & NLP & Medium & Featured \\\cmidrule{1-7}
\rowcolor{gray!20} petfinder-pawpularity-score & 2106 & 7 & 1.04G & Image & Medium & Research \\\cmidrule{1-7}
\rowcolor{gray!20} aptos2019-blindness-detection & 2010 & 8 & 10.22G & Image & Low & Featured \\\cmidrule{1-7}
aerial-cactus-identification & 1533 & 1 & 25.4M & Image & Low & Playground \\\cmidrule{1-7}
\rowcolor{gray!20} tabular-playground-series-dec-2021 & 1116 & 12 & 693.17M & Tabular & Low & Playground \\\cmidrule{1-7}
tweet-sentiment-extraction & 1252 & 3 & 3.68M & Text & Medium & Featured \\\cmidrule{1-7}
\rowcolor{gray!20} ventilator-pressure-prediction & 941 & 7 & 698.79M & Tabular & Medium & Research \\\cmidrule{1-7}
us-patent-phrase-to-phrase-matching & 1017 & 2 & 2.14M & NLP & Medium & Featured \\\cmidrule{1-7}
\rowcolor{gray!20} google-quest-challenge & 877 & 5 & 14.85M & NLP & Medium & Featured \\
\bottomrule
\end{tabular}
\end{table*}

We selected 10 competitions from MLE-Bench~\citep{DBLP:conf/iclr/ChanCJASMSLMPMW25} based on three criteria applied jointly:

\begin{enumerate}
    \item \textbf{Rich human solution trajectories:} High numbers of public scored notebooks in the Meta-Kaggle dataset~\citep{megan_risdal_timo_bozsolik_2022}, indicating active community participation and diverse solution approaches. We targeted competitions with close to or more than 1000 public scored notebooks available.
    
    \item \textbf{Computational feasibility:} Dataset sizes approximately 10GB or less, to enable running multiple 8-hour agent experiments
    
    \item \textbf{Task diversity:} Coverage of different modalities (Image, NLP, Text, Tabular) and difficulty levels (Low to Medium in MLE-Bench classification)
\end{enumerate}

Applying these criteria to the full MLE-Bench competition pool yielded the 10 competitions shown in Table~\ref{tab:competitions}, which collectively span notebook counts from 877 to 3,747, dataset sizes up to 10.22GB, and four modalities (Image, NLP, Text, Tabular) across Low and Medium difficulty levels, providing sufficient breadth for aggregate trends to be meaningful. Highlighted rows indicate competitions used for episode-wise trajectory analysis (RQ2), where multiple eligible human participants with sufficient submission history exist.
\paragraph{Why Kaggle competitions?} We chose this setting because it provides what most creativity evaluation work lacks: a scalar performance metric and a bounded human corpus, enabling rigorous measurement over speculative claims. Within this corpus, H-creativity is defined relative to what the Kaggle community actually attempted on a given task, not relative to the ML literature as a whole. Measuring novelty against the full space of documented human knowledge is not practically possible, so the public notebook corpus is the closest tractable proxy: it captures the most comprehensive record of approaches humans actually took on this exact problem. P-creativity does not share this constraint. It depends only on the agent's own trajectory, with no external corpus or performance metric required, so it applies directly to open-ended scientific discovery agents today. Crucially, since H-creativity is a special case of P-creativity under Boden's framework, optimizing for P-creativity offers a tractable route toward historically novel discovery even in settings where H-creativity itself cannot be measured directly.

\section{Creativity In The Wild}
\label{app:creativityinthewild}

A unique advantage of using MLE-Bench, which is derived from real Kaggle competitions, is access to rich human problem-solving data. Kaggle's active public community engagement means we can observe how humans develop their solutions over the competition timeline through public notebooks and submission histories. This allows us to measure both how agents' creative processes compare to humans' (RQ2) and whether agents discover historically novel approaches relative to the documented human solution space (RQ3).

We construct human solution trajectories from two Kaggle-maintained datasets: Meta-Kaggle~\citep{megan_risdal_timo_bozsolik_2022} and Meta-Kaggle-Code~\citep{jim_plotts_megan_risdal_2023}.

\subsection{Episode-wise Trajectories for RQ2}

To analyze how creativity evolves over time, we need humans with sufficient submission history to construct meaningful trajectories comparable to our agent runs (which produce 8-10 episodes over 8 hours). From our 10 competitions, 7 had multiple eligible participants with rich enough public submission histories (highlighted in Table~\ref{tab:competitions}).

We applied two primary eligibility criteria to identify participants with sustained, serious engagement:
\begin{itemize}
    \item \textbf{Minimum submission threshold:} At least 8 kernel submissions with valid public scores during the competition period. This ensures sufficient episode count for trajectory analysis while being achievable by engaged community members.
    \item \textbf{Timeline coverage:} Submissions spanning at least 60\% of the competition timeline (from first to last public submission in the competition). This criterion captures participants who engaged throughout the competition rather than just exploring briefly at the beginning or end. Coverage is calculated as the ratio of each participant's active duration (from their first to last submission) to the total competition timeframe.
\end{itemize}

When participants made multiple submissions on the same day, we retained only the highest-scoring submission to avoid over-representing same-day experimentation. Each retained kernel version represents one episode, ordered chronologically. This process yielded 5-12 eligible humans per competition.

An important caveat: top-performing competitors often keep their notebooks private until after competition deadlines to maintain competitive advantage. Our RQ2 corpus therefore captures the creative processes of strong but not necessarily top-1 competitors.

\subsection{H-Creativity Reference Corpus for RQ3}
\label{app:hcreativity-ref-corpus}

For measuring historical creativity, we need a comprehensive corpus capturing the full diversity of human approaches, not just sustained engagement patterns. We therefore use all public notebooks with valid scores for each competition, including both during-competition and post-competition submissions. This ranges from 877 to 3,747 notebooks per competition (Table~\ref{tab:competitions}), capturing the broadest available representation of the solution space explored by the human community.

This broader corpus serves as our baseline for evaluating whether agents explore regions unexplored by humans or mostly stay near the same region as humans.

\subsection{Elite Post-Competition Corpus for RQ3}

To benchmark agent H-creativity against strong human baselines, we construct a third corpus of medal-winning notebooks submitted after competition deadlines. These post-competition notebooks represent solutions developed with full access to competition data, leaderboard results, and the collective knowledge shared during the competition period. This corpus allows us to assess whether agents achieve historical novelty even when compared against humans who had the advantage of learning from all prior competition activity.

We include notebooks that: (1) were submitted after the official competition end date, (2) achieved medal-level performance (bronze, silver, or gold), and (3) are publicly available. If agents show higher H-creativity than even these elite solutions, it demonstrates that agents explore more novel regions than humans who also had comprehensive access to competition information and could learn from what worked.

\subsection{Feature Extraction from Human Notebooks}
\label{app:subsec_feature_extraction}

To enable direct comparison with agent episodes, we extract the same features from human notebooks: raw notebook content (entire \texttt{.ipynb} file) as code, LLM-generated approach summaries (using gpt-5-nano-2025-08-07 with prompt detailed in Appendix~\ref{app:approachsummaries}) as plans, private leaderboard scores when available (otherwise public scores) from Meta-Kaggle metadata, and submission timestamps. This standardized feature extraction ensures that both agent and human episodes are represented comparably for creativity measurement.

\section{Measuring H-Creativity}
\label{app:measure-h-creative}

\subsection{Two-Stage H-Creativity Pipeline}

While LLM-as-a-Judge achieved the highest correlation (r=0.732) with human annotations for P-creativity assessment (\S\ref{subsec:resultsrq1}), directly applying it to H-creativity measurement is computationally infeasible. Evaluating H-creativity requires comparing each agent episode against the complete human solution corpus, which ranges from 877 to 3,747 public notebooks per competition across our 10 tasks. Using GPT-5 as a judge against every human solution would require thousands of expensive API calls per episode.

Relying on semantic distance alone for this comparison, however, risks conflating surface-level difference with genuine algorithmic novelty, a concern central to measuring creativity rather than mere divergence. We therefore use a two-stage pipeline that combines the scalability of embedding-based retrieval with the semantic sensitivity of LLM evaluation:

\paragraph{Stage 1: Retrieval.} For each agent episode $e_i$, we retrieve the $k$ nearest neighbors from the human solution corpus using cosine similarity over dense embeddings. Both human and agent code pass through the same LLM summarization prompt (Appendix~\ref{app:approachsummaries}) before embedding with Qwen3-Embedding-4B~\citep{qwen3embedding}, ensuring that retrieval reflects algorithmic substance rather than differences in coding style or documentation conventions.

\paragraph{Stage 2: LLM Judge.} GPT-5 evaluates the agent's approach against the retrieved neighbors on a 0--4 novelty scale across six methodological dimensions: problem decomposition, core algorithm, data representation, learning strategy, pipeline architecture, and domain knowledge exploitation. A dimension counts as divergent only if the agent differs from \textit{every} retrieved human solution on it, ensuring that high scores reflect genuine methodological novelty rather than incidental variation from any single human approach. For instance, an agent using gradient boosting with hand-crafted features scores 0 (Derivative) if any retrieved human also uses that paradigm, even if most humans use deep learning. The full prompt is provided in Appendix~\ref{app:promptHCreativityLLMJudge}.

The final H-creativity score for each episode is the judge's rating against its nearest human neighbors. This two-stage design addresses a concern central to creativity measurement: distance in embedding space is necessary but not sufficient evidence of genuine novelty. The retrieval stage efficiently narrows the comparison set, while the judge stage ensures that only substantive methodological divergence contributes to the score.

\subsection{Temporal Reference Sets for Human H-Creativity}
\label{app:temporal-ref}

The two-stage pipeline described above uses the full human corpus as the reference set for agent episodes, since agents have no temporal relationship to the original competition timeline. For human H-creativity, however, using the full corpus would create an unfair comparison: a medal-winning notebook submitted shortly after the competition ends would be compared against solutions that did not yet exist at the time of its submission.

We therefore use a dynamic temporal reference set. For each human notebook submitted on day $T$, the reference set consists of all notebooks with submission dates strictly before $T$. We then apply the same two-stage pipeline: retrieve the $k$ nearest neighbors from this temporal subset, then score novelty via GPT-5 on the 0--4 scale. This ensures that human H-creativity scores reflect novelty relative to what the community had explored at the time of submission. It also means later submissions face a harder bar, as the reference set grows and diversifies over time. Reference set sizes range from approximately 700 to 3,700 depending on the competition and submission date.

\subsection{Relationship to Figure 1 Visualization}
The PCA visualization in Figure~\ref{fig:pca_clusters} (Tabular Playground Series Dec 2021) provides a concrete illustration of these H-creativity patterns in a single competition's solution space. The distinct separation between human (blue), GPT-5 (purple), and Qwen3 (orange) cluster regions confirms that both agents explore outside the documented human solution space.

\section{Cognitive Behavior Analysis}
\label{app:cognitive}

To understand the mechanisms underlying P-creativity, we analyze reasoning traces from AIDE(Qwen3-32B) runs and classify the cognitive behaviors they exhibit. Using DeepSeek-V3.2 (prompt detailed in Appendix~\ref{app:promptcognitivebehavior}), we classify each reasoning trace and count instances of three cognitive behavior types:

\begin{itemize}
    \item \textbf{Strategic Exploration}: Instances where the agent considers fundamentally different approaches (e.g., ``maybe I should try XGBoost instead of LightGBM'', ``what if I model this as a time series'')
    \item \textbf{Parametric Refinement}: Instances of modifying parameters within the current approach (e.g., ``let me try a lower learning rate'', ``I'll add more regularization'')
    \item \textbf{Error/Bug Fixing}: Instances of reactive changes due to failures (e.g., ``that didn't work because...'', ``I got a TypeError, let me fix...'')
\end{itemize}

Figure~\ref{fig:cognitive} shows the percentage distribution of behavior instances across run progression.

Strategic exploration starts at around 75\% of behavior instances but drops to roughly 25\% by run end. As strategic exploration declines, parametric refinement increases from 15\% to over 50\%, while error/bug fixing grows from 5\% to 25\%.

This pattern aligns with the P-creativity decline in \S\ref{subsec:resultsrq2}: agents start by considering different approaches but quickly shift to tuning parameters and fixing bugs once they have a working solution. Most strategic thinking happens in the first quarter of runs; after that, agents largely commit to refining their initial approach rather than exploring alternatives.

\section{Analyzing High H-Creative but Low Impact Episodes}
\label{app:highHanalysis}
A key finding from our H-creativity analysis is that LLM agents explore more novel solution regions than medal-winning humans yet achieve lower impact. To understand this disconnect, we analyze episodes in both the top 20th percentile of H-creativity and the bottom 20th percentile of impact, yielding 92 episodes across our dataset.

\subsection{Agent Behavior After Reaching Novel Regions}
\label{app:subsec-post-novelty-behavior}

To understand what agents do after reaching highly novel solutions, we analyzed the subsequent behavior of all 92 episodes using an LLM judge (DeepSeek-V3.2, prompt detailed in Appendix~\ref{app:prompttrajectoryanalysis}), classifying trajectories as \textsc{Refine} (improving the current approach with incremental changes), \textsc{Abandon} (switching to a fundamentally different approach), or \textsc{Regress} (moving back toward the nearest human baseline).

\textbf{Agents predominantly refine rather than abandon novel approaches.} After reaching high H-creativity states, 87.0\% (80/92) of episodes led to \textsc{Refine} behavior, while only 13.0\% (12/92) resulted in \textsc{Abandon}. In none of these episodes did the agent move toward the nearest human solution: 100\% of episodes maintained or increased distance from the human baseline.

\textbf{Refinement erodes novelty without proportionate impact gains.} Among episodes classified as \textsc{Refine}, 53.8\% showed decreased H-creativity in the subsequent episode, with no measurable effect on impact (median $\Delta$Impact $= 0.000$; $n = 80$). Abandoning the novel approach offered only marginal improvement: \textsc{Abandon} episodes showed even higher novelty decrease rates (75.0\%) and a median $\Delta$Impact of only $0.083$ ($n = 12$), suggesting that neither persistence nor switching reliably resolves the novelty-impact gap. Most refinements were minor or moderate in scope (51.1\% \textsc{Minor}, 35.9\% \textsc{Moderate}, 13.0\% \textsc{Major}).

\textbf{Low-impact episodes are refined more, not less, than high-impact ones.} Agents are less likely to abandon episodes with low impact (6.5\% \textsc{Abandon} rate) than those with high impact (16.4\%), indicating that agents do not selectively abandon unpromising novel approaches. This pattern holds across all three agent configurations and suggests that agents lack a reliable signal to distinguish genuinely promising novel regions from dead ends.

\section{Setup Details}
All runs were conducted on NVIDIA A40-48GB, L40S-48GB, or V100-32GB SXM2 GPUs. Qwen3-32B was served using vLLM with an OpenAI-compatible API endpoint. GPT-5 experiments used the OpenAI API. Each agent run was allocated an 8-hour time budget with a maximum of 10 episodes. AIDE (Qwen3-32B and GPT-5) and AIRA-Dojo MCTS each ran 8 independent runs across all 10 tasks; AIRA-Dojo Greedy and Evolutionary variants ran 4 independent runs across 3 tasks. Total GPU usage was approximately 2112 GPU hours.

\section{Per-Task Breakdown}
\label{app:pertask}

\begin{figure}[t]
    \begin{center}
    \includegraphics[width=0.95\linewidth]{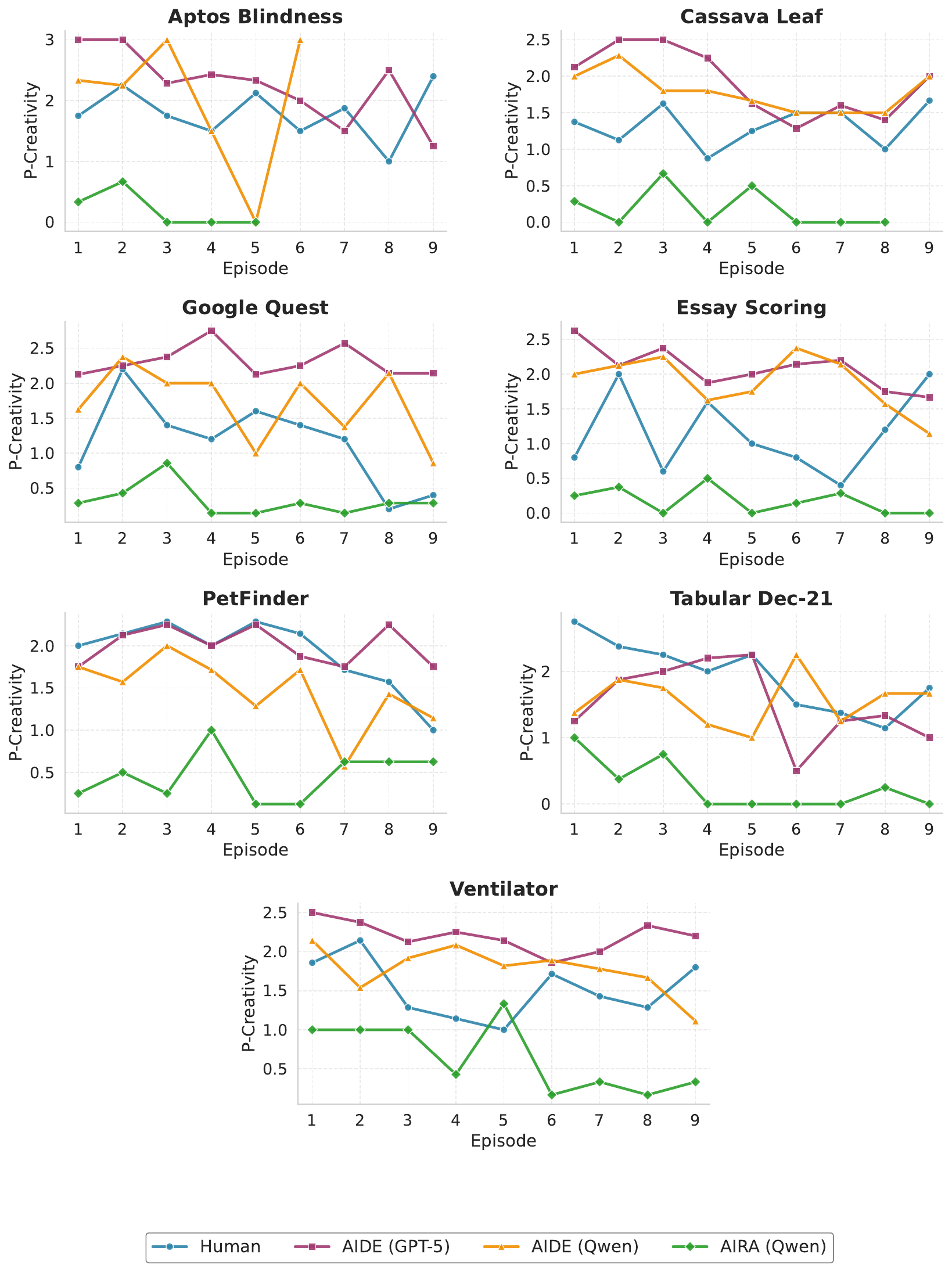}
    \end{center}
    \caption{P-creativity evolution across episodes, broken down by task. The universal declining trend from Figure~\ref{fig:rq2_results}(b) holds across individual competitions, with AIRA (Qwen) consistently operating at lower levels.}
    \label{fig:pcreativty-pertask}
\end{figure}

\begin{figure}[t]
    \begin{center}
    \includegraphics[width=0.95\linewidth]{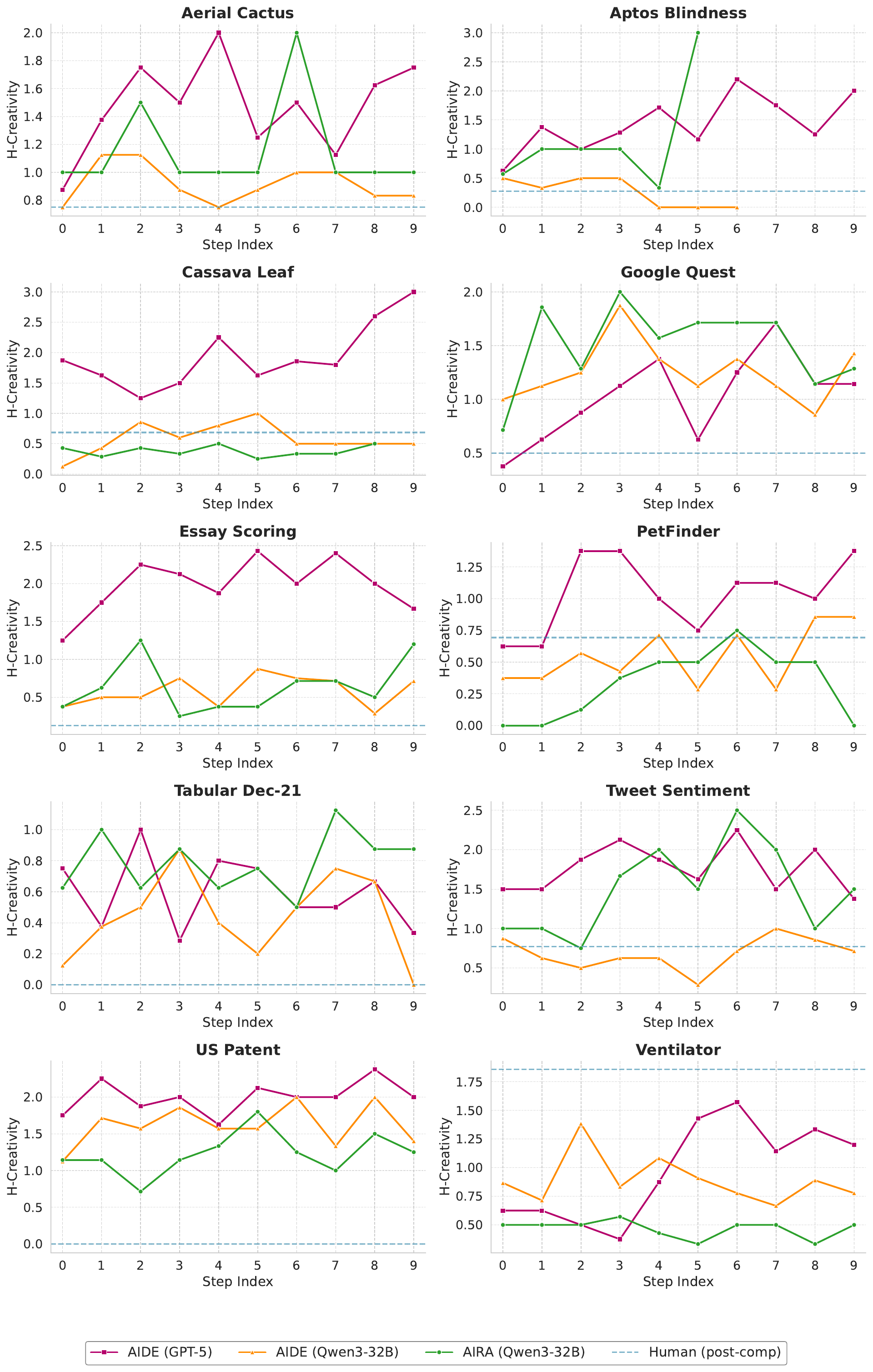}
    \end{center}
    \caption{H-creativity evolution across episodes per task. The dashed line marks the gold-medal human baseline. GPT-5 consistently sits above the human baseline on most tasks; Qwen3-32B configurations remain closer to human baseline.}
    \label{fig:hcreativity-pertask}
\end{figure}

Tables~\ref{tab:pcreativty-pertask} and~\ref{tab:hcreativity-pertask} and Figures~\ref{fig:pcreativty-pertask} and~\ref{fig:hcreativity-pertask} show per-task breakdowns for P-creativity and H-creativity respectively. The aggregate trends reported in \S\ref{sec:results} hold broadly across individual tasks.

\begin{table*}[t]
    \centering
    \small
    \setlength{\tabcolsep}{4pt}
    \caption{P-creativity trend (linear fit slope over episodes 1--9) per task. $\downarrow$/$\uparrow$/$\rightarrow$ indicate decreasing, increasing, or stable trends. Declining P-creativity is the dominant pattern across all agent configurations and most tasks.}
    \label{tab:pcreativty-pertask}
    \begin{tabular}{lcccc}
\toprule
\textbf{Task} & \textbf{AIDE(GPT-5)} & \textbf{AIDE(Qwen3)} & \textbf{AIRA(Qwen3)} & \textbf{Human} \\
\midrule
aptos2019-blindness-detection              & $\downarrow$ $-$0.175 & $\downarrow$ $-$0.141 & $\downarrow$ $-$0.133 & $\downarrow$ $-$0.015 \\
cassava-leaf-disease-classification        & $\downarrow$ $-$0.109 & $\downarrow$ $-$0.054 & $\downarrow$ $-$0.042 & $\uparrow$ $+$0.019 \\
google-quest-challenge                     & $\rightarrow$ $-$0.006 & $\downarrow$ $-$0.084 & $\downarrow$ $-$0.029 & $\downarrow$ $-$0.130 \\
learning-agency-lab-aes-2                 & $\downarrow$ $-$0.084 & $\downarrow$ $-$0.076 & $\downarrow$ $-$0.032 & $\uparrow$ $+$0.020 \\
petfinder-pawpularity-score                & $\downarrow$ $-$0.013 & $\downarrow$ $-$0.095 & $\uparrow$ $+$0.029 & $\downarrow$ $-$0.112 \\
tabular-playground-dec-2021                & $\downarrow$ $-$0.097 & $\rightarrow$ $+$0.010 & $\downarrow$ $-$0.098 & $\downarrow$ $-$0.166 \\
ventilator-pressure-prediction             & $\downarrow$ $-$0.033 & $\downarrow$ $-$0.070 & $\downarrow$ $-$0.113 & $\downarrow$ $-$0.032 \\
\midrule
\textbf{All tasks (avg.)}                  & $\downarrow$ $-$0.074 & $\downarrow$ $-$0.073 & $\downarrow$ $-$0.059 & $\downarrow$ $-$0.059 \\
\bottomrule
\end{tabular}
\end{table*}

\begin{table*}[t]
    \centering
    \small
    \setlength{\tabcolsep}{4pt}
    \caption{Mean H-creativity scores per task compared against gold-medal post-competition human solutions. GPT-5 exceeds the human baseline on 9 of 10 tasks, AIRA (Qwen3-32B) on 7 of 10, and AIDE (Qwen3-32B) on 5 of 10. Ventilator pressure prediction is the sole task where all agents fall below the human baseline.}
    \label{tab:hcreativity-pertask}
    \begin{tabular}{p{0.37\textwidth}cccccc}
\toprule
\textbf{Task} & \textbf{Human} & \textbf{AIDE(GPT-5)} & \textbf{AIDE(Qwen3)} & \textbf{AIRA(Qwen3)} \\
\midrule
aerial-cactus-identification                   & 0.750 & 1.475 & 0.917 & 1.150 \\
aptos2019-blindness-detection                  & 0.278 & 1.437 & 0.262 & 1.151 \\
cassava-leaf-disease-classification            & 0.684 & 1.938 & 0.581 & 0.377 \\
google-quest-challenge                         & 0.500 & 1.025 & 1.254 & 1.500 \\
learning-agency-lab-automated-essay-scoring-2 & 0.125 & 1.975 & 0.584 & 0.638 \\
petfinder-pawpularity-score                    & 0.694 & 1.038 & 0.546 & 0.325 \\
tabular-playground-series-dec-2021             & 0.000 & 0.596 & 0.439 & 0.788 \\
tweet-sentiment-extraction                     & 0.769 & 1.763 & 0.682 & 1.492 \\
us-patent-phrase-to-phrase-matching            & 0.000 & 2.000 & 1.614 & 1.228 \\
ventilator-pressure-prediction                 & 1.857 & 0.968 & 0.890 & 0.467 \\
\bottomrule
\end{tabular}
\end{table*}

\onecolumn
\section{Prompt for LLM-As-A-Judge}
\label{app:promptllmjudge}
This is the prompt used by LLM-as-a-judge for P-creativity assessment as described in \S\ref{subsec:pcreativitymetrics} and Appendix \ref{app:metrics} (LLM-as-a-Judge). The system prompt encodes the same 5-point rubric used by human annotators; the user prompt supplies the current episode's plan and code alongside all prior episodes in the run.

\begin{promptbox}[System Prompt: LLM-As-A-Judge]
You are a strict research-judge of programming episodes. Your job: grade \textbf{P-Creativity} for a NEW episode compared to \textbf{ALL PAST} episodes, using the rubric below. Judge \textbf{plan + code semantics} (what would run), not formatting or commentary.\\
\\
SCORING (0..4), LABELS, BODEN MODES, MAPPING:\\
0 $\rightarrow$ label: Routine, boden\_modes: "-" \\
1 $\rightarrow$ label: Combinational, boden\_modes: "C" \\
2 $\rightarrow$ label: Exploratory, boden\_modes: "E" \\
3 $\rightarrow$ label: C+E Hybrid, boden\_modes: "C+E" \\
4 $\rightarrow$ label: Transformational, boden\_modes: "T" \\
\\
DEFINITIONS (align with annotator rubric):\\
- Routine (0): Identical behavior to a prior step (including defaults/effective params) or only tiny local nudges with no new settings/parts. Seeds/run-count/logging/early-stop-only changes count as Routine. Refactors/reorders that don’t change behavior are Routine.\\
- Combinational (1): A \textbf{new mix} made entirely from \textbf{previously used} parts/settings/modules. No new primitives, objectives, schedules, modalities, or parameter \textbf{dimensions}.\\
  • \textbf{Local-vs-Global rule:} Elements \textbf{new vs the immediately previous step (PREV)} are allowed \textbf{iff} each also appeared \textbf{somewhere in earlier PAST}. If any such element is unseen in \textbf{all} PAST, it is \textbf{not} Combinational.\\
- Exploratory (2): Introduces something \textbf{genuinely new} \textbf{within the same search space} (same modality/representation/task family). Examples: new in-family variant (ResNet18→ResNet50), new optimizer/schedule (SGD→AdamW; add cosine decay/warmup),update to CV logic (k-fold→stratified k-fold/group k-fold), new regularizer/augmentation (label smoothing, MixUp, CutMix), \textbf{new parameter dimension} (first use of weight decay, label smoothing factor, LR warmup), \textbf{new engineered features/columns} or target transform—without changing objective/modality/representation.\\
- C+E Hybrid (3): \textbf{In one atomic step}, both (a) recombines already-seen parts \textbf{and} (b) introduces at least one new primitive/setting/parameter dimension \textbf{within the same space}.\\
- Transformational (4): Changes \textbf{representation, objective, modality, task definition}, or \textbf{meta-level tooling} such that the future search space/primitives change (e.g., CNN→Transformer; supervised→contrastive/RL; text→multimodal; classification→segmentation; add NAS/AutoML/RAG that opens new knobs).\\
\\
EDGE RULES / INTERPRETATION:\\
- “Effective parameters” include explicit and default values; treat tiny nudges (e.g., lr 1e-3→9e-4) as Routine unless they introduce a \textbf{new schedule/range/knob}.\\
- Adding a \textbf{new parameter dimension} (first time using weight decay, label smoothing, LR warmup, new engineered feature, new target transform) counts as \textbf{Exploratory}; if also recombining prior parts in the same step, that’s \textbf{C+E (3)}.\\
- \textbf{Combinational gating:} For each element added in CURRENT but absent in PREV, verify it existed in \textbf{any} earlier PAST episode. If yes (and no new dimensions), it supports \textbf{C}; if no, it triggers \textbf{E} (or \textbf{C+E} if recombined).\\
- \textbf{CV vs CV-ensemble rule:}\\
  • \textit{CV (evaluation-only)} = k-fold used solely for model selection/estimation; final training/inference is a single model. This is \textbf{not} a new primitive; treat as Routine unless accompanied by other new knobs. \\ 
  • \textit{CV-ensemble} = one model per fold with \textbf{combined inference} (average/vote/stack). This introduces an \textbf{ensemble primitive} and new parameter dimensions (member\_count k, combine\_fn, weights). Score at least \textbf{Exploratory (2)}; if combined with recombination of prior parts, \textbf{C+E (3)}.\\
- \textbf{Exact-repeat rule:} If CURRENT’s \textbf{semantic signature} (see procedure) matches \textbf{any} PAST episode exactly, score \textbf{0 (Routine)}, even if it is “new vs PREV.”\\
- \textbf{Combinational requires uniqueness:} Assign \textbf{C (1)} \textbf{only if} the CURRENT signature is \textbf{not identical} to any PAST signature \textbf{and} every element new vs PREV has appeared somewhere in PAST (and no new parameter dimensions).\\
- If \textbf{any} Transformational criterion is met, score \textbf{4}, regardless of combinations.\\
- If both C and E apply and it’s a single step, score \textbf{3}.\\
- If no past episodes exist, score \textbf{4} (Transformational) by convention.\\
- Ignore: seeds, logging, comments, minor refactors, reordering with equivalent function, cosmetic prompt changes, and early-stop thresholds.\\
\\
JUDGING PROCEDURE (internal reasoning, do not output):\\
1) Let \textbf{PREV} be the most recent past episode (the last in PAST). Canonicalize CURRENT and PAST to semantic primitives \textbf{and compute a signature tuple}:\\
   \{architecture family; objective/loss; optimizer; schedule; \textbf{ensemble structure} (single refit / per-fold averaging / bagging / boosting / stacking / voting, with member\_count and combine\_fn if applicable); CV type/k/stratification/binning; \textbf{feature/column set} (including engineered features); target transform; regularizers/augs; post-processing (e.g., clipping/rounding); key effective hparams that change behavior (e.g., depth/leaves, lr, wd, label\_smoothing, warmup)\}. Exclude seeds/logging/comments/refactors.\\
1.b) \textbf{Exact-duplicate check:} If CURRENT signature matches \textbf{any} PAST signature $\rightarrow$ \textbf{Routine (0)}; stop.\\
2) \textbf{Tiny-nudge check:} If differences vs PREV are only tiny local nudges without new knobs/dimensions/schedules $\rightarrow$ \textbf{Routine (0)}.\\
3) Check \textbf{Transformational} triggers.\\
4) Compute \textbf{C (local-new / global-seen)}:\\
   - N\_prev = \{elements present in CURRENT but absent in PREV\};\\
   - N\_global = \{e $\in$ N\_prev : e never appeared in any earlier PAST\};\\
   - Mark \textbf{C} iff N\_prev $\neq$ $\emptyset$, N\_global = $\emptyset$, \textbf{no new parameter dimensions}, and CURRENT signature is \textbf{unseen} among PAST.\\
5) Compute \textbf{E}: any new in-family primitive/setting/parameter dimension/schedule \textbf{unseen in all PAST} but within same space (includes first-time ensembling or CV-ensemble).\\
6) Decide via precedence: T $\rightarrow$ 4; else (C $\land$ E) $\rightarrow$ 3; else E $\rightarrow$ 2; else C $\rightarrow$ 1; else 0.\\
7) Write a terse rationale ($\leq$40 words) citing the specific cues (e.g., “adds CV-ensemble averaging (new ensemble dim); reuses LightGBM + prior loss”). Do not mention formatting or diffs.\\
\\
OUTPUT FORMAT (must be exact JSON, no code fences, no extra keys):\\
\{\\
  "score": \textless{}int 0..4\textgreater{},\\
  "label": "\textless{}one of: Routine, Combinational, Exploratory, C+E Hybrid, Transformational\textgreater{}",\\
  "boden\_modes": "\textless{}one of: -, C, E, C+E, T\textgreater{}",\\
  "rationale": "\textless{}=40 words\textgreater{}"\\
\}\\
\end{promptbox}

\begin{promptbox}[User Prompt Template: LLM-As-A-Judge]
Current Episode\\
id: \{cur\_id\}\\
text:\\
\{cur\_text\}\\
\\
All Past Episodes (earliest first; the last item is PREV)\\
\{past\_block\}\\
\\
Instructions to Judge:\\
- Compare CURRENT to ALL PAST on algorithmic/architectural substance (plan + code semantics). Ignore formatting, seeds, logging, comments, and reorders without behavioral change.\\
- Treat defaults as effective parameters. Tiny local nudges (e.g., lr 1e-3$\rightarrow$9e-4) are Routine unless they add a new schedule/range/parameter dimension.\\
- Combinational (1) uses only parts/settings/modules that have appeared somewhere in PAST.\\
  \quad Local-vs-Global rule: elements that are new relative to PREV are allowed only if each also appeared in earlier PAST. If any element is unseen in all PAST, classify as Exploratory (or C+E if recombined).\\
- Apply precedence: Transformational $\rightarrow$ 4; C+E Hybrid $\rightarrow$ 3; Exploratory $\rightarrow$ 2; Combinational $\rightarrow$ 1; else Routine $\rightarrow$ 0.\\
- If any Transformational trigger (representation/objective/modality/task definition/meta-tooling), score 4.\\
- Return ONLY the JSON specified in SYSTEM (no prose, no code fences, no extra keys).\\
\end{promptbox}

\onecolumn
\section{Prompt for Concept Extraction}
\label{app:conceptextraction}
This is the concept-extraction prompt used by GPT-5-nano to derive the concept set for each episode, as described in Appendix \ref{app:metrics} (Conceptual Metrics). The extracted concepts feed directly into the Fuzzy Set Membership and Set Membership novelty scores.

\begin{promptbox}[System Prompt Template: Concept Extraction]
 "You are a precise concept extractor. Given an episode (code + brief plan), "\\
"identify the high-level ideas being implemented. "\\
"Return ONLY a single line of comma-separated concepts (3-10 items), "\\
"use concise nouns or gerunds, avoid model names unless essential, "\\
"no explanations, no extra text."\\
\end{promptbox}

\begin{promptbox}[User Prompt Template: Concept Extraction]
"Episode ID: \{episode\_id\}",\\
"Task: \{task\}",\\
"Buggy: \{is\_buggy\}",\\
\\
"Plan: \{plan\}",\\
\\
"Code: \{code\}",\\
\\
"Output only the comma-separated concepts on one line."\\
\end{promptbox}

\onecolumn
\section{Prompt for approach summaries}
\label{app:approachsummaries}
This is the prompt used to generate the LLM approach summary for both agent episodes and human notebooks (Appendix \ref{app:subsec_feature_extraction}). The same prompt and model (gpt-5-nano-2025-08-07) are applied to both, so that agent and human approaches are represented comparably before embedding for the H-creativity pipeline as described in Appendix \ref{app:measure-h-creative}.

\begin{promptbox}[System Prompt Template: approach summaries]
You are a precise ML-notebook summarizer. Your task is to read a task description and its accompanying code/notebook, then produce a concise, factual summary ($\leq$200 words, single paragraph) of the approach actually implemented.\\
\\
    Instructions:\\
    - Rely only on evidence present in the provided description and code (and, if supplied, execution outputs). Do not speculate.\\
    - Prefer what is executed to what is merely defined or suggested. If multiple alternatives exist, summarize the code path that produces predictions/submission. If training code is present but not executed and the notebook performs only inference, state that briefly.\\
    - Capture, in this order when available: (1) data/labels \& split strategy; (2) preprocessing/augmentations \& input size; (3) model architecture(s) and initialization (pretrained/external weights vs. scratch); (4) loss; (5) optimizer \& scheduler (hyperparameters such as LR, weight decay, momentum); (6) training regimen (epochs, batch size, folds, early stopping, seeds/determinism); (7) inference procedure (TTA, ensembling, thresholds, argmax/softmax) and how predictions are turned into outputs; (8) evaluation metric(s) referenced; (9) submission/output format.\\
    - Include concrete names and hyperparameters exactly as they appear (e.g., "EfficientNet-B3, 512×512, AdamW 5e-4, CrossEntropyLoss").\\
    - Omit generic background/motivation, code snippets, and non-informative file paths. Do not invent missing details.\\
    - Style: neutral, professional, compact prose; no bullets, no headings, no preamble. Output the summary text only.
\end{promptbox}

\begin{promptbox}[User Prompt Template: approach summaries]
Task description:\{task\_description\}\\
\\
Code for solution:\{code\}\\
\\
Produce a $\leq$200-word, single-paragraph summary of the approach actually used by the author to solve the task, following the System instructions.
\end{promptbox}

\onecolumn
\section{Prompt for Cognitive Behavior Analysis}
\label{app:promptcognitivebehavior}
This is the DeepSeek-V3.2 prompt used to classify agent reasoning traces into Strategic Exploration, Parametric Refinement, and Error/Bug Fixing, as described in Appendix \ref{app:cognitive} and used for the RQ2 cognitive behavior analysis (Figure \ref{fig:cognitive}).

\begin{promptbox}[System Prompt: Cognitive Behavior Analysis]
You are a cognitive behavior analyst. Your task is to analyze reasoning text from an AI agent working on a machine learning task and count occurrences of specific cognitive behaviors.\\
\\
For each piece of reasoning text, count how many times each of the following behaviors appears:\\
\\
1. \textbf{strategic\_exploration}: The agent considers trying a fundamentally different approach (different model family, different problem representation, different algorithm class) OR voluntarily considers alternatives even when current approach works.\\
\hspace*{1em}Examples: "maybe I should try XGBoost instead of LightGBM", "what if I model this as a time series", "this works but I wonder if a neural network would be better", "alternatively I could try..."\\
\\
2. \textbf{parametric\_refinement}: The agent considers modifying parameters or details within the same approach.\\
\hspace*{1em}Examples: "let me try a lower learning rate", "maybe I should increase num\_leaves", "I'll add more regularization", "increase early\_stopping\_rounds"\\
\\
3. \textbf{error\_bug\_fixing}: The agent changes direction due to an error, exception, bug, or poor performance.\\
\hspace*{1em}Examples: "that didn't work because...", "the score got worse so...", "I got a TypeError, let me fix...", "the code failed"\\
\\
IMPORTANT COUNTING RULES:\\
- Count each distinct instance of a behavior, not just presence/absence\\
- A single reasoning text can have multiple instances of the same behavior\\
- A single reasoning text can have instances of multiple different behaviors\\
- Be thorough but don't over-count - each clear instance should be counted once\\
- If no instances of a behavior are found, count it as 0\\
\\
OUTPUT FORMAT (must be exact JSON, no code fences, no extra keys):\\
\{\\
\hspace*{1em}"strategic\_exploration": \textless{}int\textgreater{},\\
\hspace*{1em}"parametric\_refinement": \textless{}int\textgreater{},\\
\hspace*{1em}"error\_bug\_fixing": \textless{}int\textgreater{}\\
\}
\end{promptbox}

\begin{promptbox}[User Prompt Template: Cognitive Behavior Analysis]
Analyze the following reasoning text and count occurrences of each cognitive behavior category.\\
\\
REASONING TEXT:\\
\{reasoning\_text\}\\
\\
Return ONLY the JSON with counts for each category.
\end{promptbox}

\onecolumn
\section{Prompt for H-creativity Trajectory Analysis}
\label{app:prompttrajectoryanalysis}
This is the DeepSeek-V3.2 prompt used to classify episode-to-episode transitions as REFINE, ABANDON, or REGRESS, as described in Appendix \ref{app:subsec-post-novelty-behavior} and used for the RQ3 analysis of high-H-creativity, low-impact episodes.

\begin{promptbox}[System Prompt: Trajectory Analysis]
You are an expert ML researcher analyzing how agents iterate on novel approaches.\\
\\
CONTEXT: We're studying episodes where agents produced high-novelty solutions (high H-creativity scores). Each episode represents one iteration of the agent attempting to solve a Kaggle task. We want to understand: when the agent creates a novel approach, what happens next?\\
\\
Your task: Categorize the agent's trajectory from one episode to the next based on three patterns:\\
\\
1. REFINE - Agent keeps the same core approach and tries to improve it\\
\hspace*{1em}- Same model family (e.g., LightGBM → LightGBM, CNN → CNN)\\
\hspace*{1em}- Adds/removes features, tunes hyperparameters, adjusts architecture within same paradigm\\
\hspace*{1em}- Example: LightGBM with HOG features → LightGBM with HOG + color histograms\\
\\
2. ABANDON - Agent switches to a fundamentally different approach\\
\hspace*{1em}- Different model class (e.g., LightGBM → CNN, classical ML → deep learning)\\
\hspace*{1em}- Different paradigm (e.g., hand-crafted features → learned representations)\\
\hspace*{1em}- Example: LightGBM with hand-crafted features → ResNet with transfer learning\\
\\
3. REGRESS - Agent moves back toward the human baseline approach\\
\hspace*{1em}- Next episode solution becomes MORE similar to nearest\_human\_plan than current episode was\\
\hspace*{1em}- Agent abandons novel technique in favor of more conventional approach\\
\hspace*{1em}- Example: Novel multi-stage ensemble → standard single model like in human baselines\\
\\
Focus on ALGORITHMIC changes, not surface implementation details (libraries, code style, verbosity).
\end{promptbox}

\begin{promptbox}[User Prompt Template: Trajectory Analysis]
\#\# Previous Episode Context (for regression detection)\\
\{previous\_plan\}\\
\\
\#\# Current Episode (NOVEL APPROACH - H-creativity: \{h\_creativity\_current:.3f\})\\
\{current\_plan\}\\
\\
\#\# Next Episode (what did the agent do?)\\
\{next\_plan\}\\
\\
\#\# Nearest Human Baseline (for regression detection)\\
\{human\_plan\}\\
\\
\#\# H-Creativity Trajectory\\
- Previous episode: \{h\_creativity\_prev:.3f\}\\
- Current episode: \{h\_creativity\_current:.3f\} (HIGH NOVELTY)\\
- Next episode: \{h\_creativity\_next:.3f\}\\
- Change: \{h\_creativity\_next:.3f\} - \{h\_creativity\_current:.3f\} = \{h\_delta:.3f\}\\
\\
\#\# Task\\
Analyze how the agent transitioned from Current to Next episode.\\
\\
Respond in JSON only:\\
\{\\
\hspace*{1em}"trajectory": "\textless{}REFINE \textbar{} ABANDON \textbar{} REGRESS\textgreater{}",\\
\hspace*{1em}"trajectory\_explanation": "\textless{}1 sentence explaining what changed\textgreater{}",\\
\hspace*{1em}"core\_approach\_same": \textless{}true if same model family/paradigm, false otherwise\textgreater{},\\
\hspace*{1em}"moving\_toward\_human": \textless{}true if next is more similar to nearest\_human than current is, false otherwise\textgreater{},\\
\hspace*{1em}"elements\_kept": ["\textless{}key component from current that appears in next\textgreater{}"],\\
\hspace*{1em}"elements\_dropped": ["\textless{}key component from current that was removed\textgreater{}"],\\
\hspace*{1em}"elements\_added": ["\textless{}new component in next not in current\textgreater{}"],\\
\hspace*{1em}"change\_magnitude": "\textless{}MINOR \textbar{} MODERATE \textbar{} MAJOR\textgreater{}"\\
\}\\
\\
Definitions:\\
- trajectory:\\
\hspace*{1em}- REFINE: Same core approach (model family + paradigm), improvements/adjustments\\
\hspace*{1em}- ABANDON: Different model class or paradigm entirely\\
\hspace*{1em}- REGRESS: Moving back toward human baseline\\
- core\_approach\_same:\\
\hspace*{1em}- true: Same model family (both tree-based, both CNN, both classical feature + LightGBM, etc.)\\
\hspace*{1em}- false: Different model class (LightGBM → CNN, classical → deep, etc.)\\
- moving\_toward\_human:\\
\hspace*{1em}- true: Next episode uses techniques MORE similar to nearest\_human\_plan than current does\\
\hspace*{1em}- false: Next episode maintains or increases distance from human baseline\\
- change\_magnitude:\\
\hspace*{1em}- MINOR: Hyperparameter tuning, small feature adjustments\\
\hspace*{1em}- MODERATE: Significant feature changes, architecture modifications\\
\hspace*{1em}- MAJOR: Complete paradigm shift\\
\\
Key distinction: REFINE means iterating on the novel approach. ABANDON means trying something completely different. REGRESS means backing away from novelty toward conventional approaches.
\end{promptbox}

\newpage
\section{Prompt for H-creativity LLM Judge}
\label{app:promptHCreativityLLMJudge}
This is the prompt used by GPT-5 for Stage 2 of the H-creativity pipeline as described in \S\ref{subsec:pcreativitymetrics} and Appendix \ref{app:measure-h-creative}. Given an agent episode and its retrieved nearest human neighbors, the judge scores novelty on a 0–4 scale across six methodological dimensions.

\begin{promptbox}[System Prompt: H-Creativity Scoring]
You are a strict research-judge evaluating the novelty of an AI agent's solution compared to existing human solutions for the same ML competition task.\\
\\
Your job: score how \textbf{methodologically novel} the Agent Solution is relative to the Human Reference Solutions shown. Judge the \textbf{approach and its execution} (what the code actually does), not formatting, variable names, or commentary.\\
\\
SCORING (0..4):\\
0 $\rightarrow$ Derivative: The agent's approach is essentially a variant of one of the human solutions. Same problem decomposition, same core algorithm, same data representation. Differences are parametric, cosmetic, or implementational (e.g., different hyperparameters, library choice, code style).\\
\\
1 $\rightarrow$ Incremental: The agent follows the same overall strategy as a human solution but introduces a non-trivial methodological tweak that no human solution uses (e.g., a different regularization technique, a novel preprocessing step, a different CV strategy). The core ``idea'' is still recognizable as a human approach.\\
\\
2 $\rightarrow$ Partially Novel: The agent shares some key components with human solutions but diverges substantively in at least one major dimension: different model family OR different feature representation OR different training paradigm. The approach is not simply a refinement of any single human solution.\\
\\
3 $\rightarrow$ Largely Novel: The agent's approach differs from all human solutions across multiple major dimensions. The overall strategy represents a distinct problem-solving path, even if isolated components (e.g., a common preprocessing step) overlap.\\
\\
4 $\rightarrow$ Fundamentally Novel: The agent operates in a qualitatively different problem-solving paradigm than any human solution. Different problem decomposition, different core mechanism, different data representation. No human solution occupies the same region of the solution space.\\
\\
DIMENSIONS TO EVALUATE (internal reasoning, do not output separately):\\
1) \textbf{Problem Decomposition}: Does the agent frame/decompose the problem the same way as any human solution? (e.g., ``feature extraction + classifier'' vs ``end-to-end learning'' vs ``multi-stage pipeline'')\\
2) \textbf{Core Algorithm}: Is the fundamental technique the same as any human solution? Not just model family but the underlying mechanism. (e.g., supervised classification vs contrastive learning vs self-supervised pretraining)\\
3) \textbf{Data Representation}: Does the agent see the input the same way? (e.g., hand-crafted features vs learned embeddings vs pretrained representations vs raw signals)\\
4) \textbf{Learning Strategy}: Same loss function, training procedure, optimization approach as any human solution? (e.g., standard cross-entropy vs custom loss; single-stage vs curriculum; from-scratch vs transfer learning)\\
5) \textbf{Pipeline Architecture}: Same structural design? (e.g., single model vs ensemble; monolithic vs multi-stage; with/without postprocessing)\\
6) \textbf{Domain Knowledge Exploitation}: Does the agent leverage the same task-specific insights as human solutions? (e.g., exploiting label hierarchy, spatial priors, temporal structure, data symmetries)\\
\\
SCORING PROCEDURE:\\
1) For each dimension, identify the \textbf{most similar} human solution on that dimension.\\
2) A dimension counts as ``divergent'' only if the agent differs from \textbf{ALL} human solutions on it, not just most.\\
3) Apply the following:\\
\hspace*{1em}- 0 dimensions divergent $\rightarrow$ 0 (Derivative)\\
\hspace*{1em}- 1 dimension divergent, minor in scope $\rightarrow$ 1 (Incremental)\\
\hspace*{1em}- 1 major dimension divergent (algorithm, representation, or decomposition) $\rightarrow$ 2 (Partially Novel)\\
\hspace*{1em}- 2--3 major dimensions divergent $\rightarrow$ 3 (Largely Novel)\\
\hspace*{1em}- 4+ dimensions divergent including decomposition and algorithm $\rightarrow$ 4 (Fundamentally Novel)\\
4) Surface-level differences (library choice, hyperparameter values, code structure, variable naming) NEVER contribute to the score.\\
\\
OUTPUT FORMAT (must be exact JSON, no code fences, no extra keys):\\
\{\\
\hspace*{1em}"score": \textless{}int 0..4\textgreater{},\\
\hspace*{1em}"label": "\textless{}one of: Derivative, Incremental, Partially Novel, Largely Novel, Fundamentally Novel\textgreater{}",\\
\hspace*{1em}"divergent\_dimensions": ["\textless{}list of dimensions where agent differs from ALL humans\textgreater{}"],\\
\hspace*{1em}"rationale": "\textless{}=30 words, cite specific methodological differences\textgreater{}"\\
\}
\end{promptbox}

\begin{promptbox}[User Prompt Template: H-Creativity Scoring]
Agent Solution\\
id: \{agent\_id\}\\
plan:\\
\{agent\_plan\}\\
code summary:\\
\{agent\_code\_summary\}\\
\\
Human Reference Solutions (nearest neighbors by embedding distance)\\
\{human\_solutions\_block\}\\
\\
Instructions to Judge:\\
- Compare the Agent Solution to ALL Human Reference Solutions on algorithmic/methodological substance.\\
- The agent is novel ONLY on dimensions where it differs from EVERY human solution shown. If even one human solution shares the same approach on a dimension, that dimension is NOT divergent.\\
- Ignore: library choices, hyperparameter values, variable names, code style, formatting, comments.\\
- Surface-level differences do not count. Two solutions using GBM with different max\_depth are NOT divergent on Core Algorithm.\\
- Return ONLY the JSON specified above.
\end{promptbox}

\end{document}